\documentclass{article}

\PassOptionsToPackage{square,numbers}{natbib}
\input{macros-arXiv.sty}

\usepackage[utf8]{inputenc} 
\usepackage[T1]{fontenc}    
\usepackage{hyperref}       
\usepackage{url}            
\usepackage{booktabs}       
\usepackage{amsfonts}       
\usepackage{nicefrac}       
\usepackage{microtype}      
\usepackage{xcolor}         

\usepackage{bm, amssymb}       
\usepackage{nicefrac}       
\usepackage{tikz} 
\usepackage{float}

\usepackage{array}
\usepackage{diagbox}
\usepackage{mathtools}
\usepackage{multirow, makecell}
\usepackage{rotating}
\usepackage{graphicx}
\usepackage{mathdots}

\usepackage{caption}

\usepackage{float}
\usepackage{setspace}
\usepackage{fullpage}

\usepackage[bottom]{footmisc}

\usepackage{threeparttable}

\usepackage{multirow}
\usepackage{makecell}
\usepackage{arydshln}
\usepackage{tabularx}

\usepackage{arydshln}

\usepackage{enumerate}
\usepackage{enumitem} 

\usepackage[sort, numbers]{natbib}

\usepackage{float}
\floatstyle{plain}
\newfloat{floatalgo}{ht}{floatalgo}

\usepackage{adjustbox}

\usepackage{caption}

\usepackage{mathrsfs}

\usepackage{wrapfig}

\usepackage{subfigure}
\usepackage{math_command}

\usepackage{enumitem}

\title{Accelerating Distributed Optimization: A Primal-Dual Perspective on Local Steps}

\author{
Junchi Yang 
$^\dagger$
\and
Murat Yildirim
$^\ddagger$
\and
Qiu Feng
$^\dagger$
}

\begin{document}
\maketitle
\def\thefootnote{$\dagger$}\footnotetext{Energy Systems and Infrastructure Analysis Division, Argonne National Laboratory, USA. 
Emails: \texttt{junchi.yang@anl.gov},
\texttt{fqiu@anl.gov},
}
\def\thefootnote{$\ddagger$}\footnotetext{
Industrial and Systems Engineering, Wayne State University, USA.
Emails: \texttt{murat@wayne.edu}
}
\def\thefootnote{\arabic{footnote}}

\begin{abstract}
In distributed machine learning, efficient training across multiple agents with different data distributions poses significant challenges. Even with a centralized coordinator, current algorithms that achieve optimal communication complexity typically require either large minibatches or compromise on gradient complexity. In this work, we tackle both centralized and decentralized settings across strongly convex, convex, and nonconvex objectives. We first demonstrate that a basic primal-dual method, (Accelerated) Gradient Ascent Multiple Stochastic Gradient Descent (GA-MSGD), applied to the Lagrangian of distributed optimization inherently incorporates local updates, because the inner loops of running Stochastic Gradient Descent on the primal variable require no inter-agent communication. Notably, for strongly convex objectives,  (Accelerated) GA-MSGD achieves linear convergence in communication rounds despite the Lagrangian being only linear in the dual variables. This is due to a structural property where the dual variable is confined to the span of the coupling matrix, rendering the dual problem strongly concave. When integrated with the Catalyst framework, our approach achieves nearly optimal communication complexity across various settings without the need for minibatches. 
\end{abstract}

\section{Introduction}

We consider unconstrained distributed optimization problems in the following formulation:
\begin{equation}
\label{objective}
\min _{x \in \mathbb{R}^n} F(x):=\frac{1}{M} \sum_{m=1}^M F_m(x).
\end{equation}
Here, $M$ clients, each with access to the gradient of its respective local function $F_m$, collaborate to minimize the aggregate of their objective functions. Such problems have garnered significant attention due to their relevance in diverse fields including machine learning \citep{xing2016strategies}, signal processing \citep{rabbat2004distributed}, power systems \citep{bidram2014distributed}, and control \citep{nedic2018distributed}. Particularly in Federated Learning \citep{konevcny2016federated, mcmahan2017communication}, this form is prevalent as each agent possesses unique local data, so the local function is represented by \(F_m(x) = \mathbb{E}_{\xi \sim P_m} f(x; \xi)\), where \(P_m\) denotes the data distribution for the $m$-th agent. This approach allows all agents  to train a global machine learning model represented by weight \(x\), while ensuring that no local data are directly exchanged.

\begin{table*}[t]
		\centering
		\small
\caption{Complexities for Centralized Strongly Convex Problems. This table compares communication and sample complexities in deterministic and stochastic settings, where \(L\) is the smoothness constant, \(\mu\) the strong convexity constant, and \(\kappa \triangleq L/\mu\) the condition number. \textcolor{red}{$^\star$}Only minibatch Accelerated SGD uses minibatches; all other algorithms use local steps. \textcolor{red}{$^\dagger$}ProxSkip complexities for the stochastic setting in \citep{hu2023tighter} are asymptotic, and \(\kappa\) dependency is not specified for stochastic sample complexity. NA in the ``Linear Speedup'' column indicates that the stochastic setting is not considered.}

		\renewcommand{\arraystretch}{1.49}
  \hspace*{-1em}
		\begin{threeparttable}[b]
			\begin{tabular}{c | c | c | c | c | c}
				\hline
				\hline
                     \textbf{Algorithms} &
				\textbf{Deter. Comm.} & 
				\textbf{Deter. Samp.}  & 
				\textbf{Stoc. Comm.}  &
                     \textbf{Stoc. Samp.} &
                    \textbf{Linear speedup} 
				\\
				\hline 
				\hline
				\hspace{-1mm} Minibatch Acc-SGD~\citep{woodworth2020minibatch}$\textcolor{red}{^\star}$
				& $\widetilde \cO(\sqrt{\kappa})$
				& $\widetilde \cO(\sqrt{\kappa})$
				&  $\widetilde \cO(\sqrt{\kappa})$
                    & $\widetilde \cO(\epsilon^{-1}/\mu)$ 
                    & Yes
                    \\
				\hline
		       ProxSkip~\citep{mishchenko2022proxskip, hu2023tighter}
				& $\widetilde \cO(\sqrt{\kappa})$
				& $\widetilde \cO(\kappa)$
				&  $\widetilde \cO(\sqrt{\kappa}) \textcolor{red}{^\dagger}$
                    & $\widetilde \cO(\kappa^2\epsilon^{-1}/L^2)\textcolor{red}{^\dagger}$ 
                    & No
                    \\
				\hline
				 DualFL~\citep{park2023dualfl}
				& $\widetilde \cO(\sqrt{\kappa})$
				& $\widetilde \cO(\kappa)$
				&  N/A
                    &  N/A 
                    & N/A
                    \\
                    \hline
                    
			    FGD+FSFOM \citep{sadiev2022communication}
				& $\widetilde \cO(\sqrt{\kappa})$
				& $\widetilde \cO(\kappa^{\frac{3}{4}})$
				&  N/A
                    &  N/A
                    & N/A
                    \\
				\hline
                    SCAFFOLD~\citep{karimireddy2020scaffold}
				& $\widetilde \cO(\kappa)$
				& $\widetilde \cO(\kappa)$
				&  $\widetilde \cO(\kappa)$
                    & $\widetilde \cO(\epsilon^{-1}/\mu)$ 
                    & Yes
                    \\
				\hline
                   \textcolor{blue}{Catalyst+GA-MSGD}
				& \textcolor{blue}{$\widetilde \cO(\sqrt{\kappa})$}
				& \textcolor{blue}{$\widetilde \cO(\sqrt{\kappa})$}
				&  \textcolor{blue}{$\widetilde \cO(\sqrt{\kappa})$}
                    & \textcolor{blue}{$\widetilde \cO(\kappa^{\frac{1}{2}}\epsilon^{-1}/\mu)$} 
                    & \textcolor{blue}{No}
                     \\
				\hline
                   \hspace{-2mm}\textcolor{blue}{Catalyst+SCAFFOLD}
				& \textcolor{blue}{$\widetilde \cO(\sqrt{\kappa})$}
				& \textcolor{blue}{$\widetilde \cO(\sqrt{\kappa})$}
				&  \textcolor{blue}{$\widetilde \cO(\sqrt{\kappa})$}
                    & \textcolor{blue}{$\widetilde \cO(\kappa^{\frac{1}{2}}\epsilon^{-1}/\mu)$} 
                    & \textcolor{blue}{Yes}
                    \\
				\hline
				Lower bound \citep{woodworth2020minibatch}
				& $\widetilde\Omega(\sqrt{\kappa})$
				& $\widetilde\Omega(\sqrt{\kappa})$
				&  $\widetilde\Omega (\sqrt{\kappa})$
                    & $\Omega(\epsilon^{-1}/\mu)$ 
                    & Yes
                    \\
				\hline
				\hline
			\end{tabular}
		\end{threeparttable}

\vspace{-2mm}
  \label{table:summary_sc}
 \end{table*}

The trend toward larger machine learning models and training with more extensive datasets has introduced new challenges in distributed training. The transmission of gradient or model information incurs substantial communication overhead not only due to bandwidth constraints but also because of latency and the need for privacy-preserving protocols \citep{konecny2016federated}. Additionally, when each client possesses a substantial pool of local data, gradient estimates are usually noisy. These factors challenge traditional algorithms that perform communication after each local update with exact gradients \citep{nedic2014distributed, shi2015extra, nedic2017achieving, yuan2018exact, tang2018d}. A popular method known as FedAvg~\citep{mcmahan2017communication}, or local SGD, entails performing multiple iterations of Stochastic Gradient Descent (SGD) on local objectives between communications. However, this algorithm suffers from a suboptimal convergence rate and is impacted by the level of heterogeneity, when the local functions or data can vary significantly between clients~\citep{zhao2018federated, karimireddy2020scaffold}.

In the centralized setup, Accelerated SGD with minibatches, whose size scales inversely with the target accuracy, achieve optimal communication and sample complexities in both strongly convex and general convex settings~\citep{woodworth2020minibatch}. However, employing such large minibatches is typically impractical in real-world applications. Although many algorithms replace minibatches with local (stochastic) gradient steps—termed here as local methods—they typically do not achieve these optimal complexities \citep{nemirovski2004prox, park2023dualfl, karimireddy2020scaffold}.
In strongly convex settings, with deterministic gradients, while local algorithms reach the optimal communication complexity of $\widetilde \cO(\sqrt{\kappa})$, where 
$\kappa$ denotes the condition number, they do so at the cost of suboptimal gradient complexity~\citep{mishchenko2022proxskip}; with noisy gradients, the optimal communication complexity is achieved only asymptotically \citep{hu2023tighter}.  Thus, designing local methods that can perform comparably to minibatch Accelerated SGD remains a challenging problem.

In the decentralized setup, decentralized parallel SGD (D-PSGD) with minibatch~\citep{lian2017can}, a stochastic variant of the subgradient method in~\citep{nedic2009distributed}, and decentralized local SGD~\citep{koloskova2020unified}, a decentralized counterpart to FedAvg, still suffer from performance degradation due to data heterogeneity and suboptimal convergence rates. 
Communication complexity lower bounds have been established at $\Omega(\kappa^{\frac{1}{2}}(1-\sigma_2)^{-\frac{1}{2}})$ for strongly convex objectives and $\Omega((1-\sigma_2)^{-\frac{1}{2}}\epsilon^{-2})$ for convex objectives~\citep{scaman2017optimal, scaman2019optimal}, where \(\sigma_2\) is the second largest singular value of the weight matrix. 
Recently, stochastic algorithms with local steps achieved complexities of \(\widetilde \cO(\kappa^2(1-\sigma_2)^{-1})\) and \(\cO((1-\sigma_2)^{-1}\epsilon^{-1})\) for strongly convex and convex objectives, respectively~\citep{alghunaim2024local}, indicating a gap with the lower bounds. In the nonconvex setting, the optimal communication complexity is $\Omega((1-\sigma_2)^{-\frac{1}{2}}\epsilon^{-2})$ and is achieved by a stochastic algorithm with minibatches~\citep{lu2021optimal}. However, a local method attains a suboptimal complexity of $\cO((1-\sigma_2)^{-1}\epsilon^{-2})$~\citep{alghunaim2024local}. This raises questions about whether local methods can reach optimal communication complexities in stochastic decentralized settings.


 \begin{table*}[t]
		\centering
		\small
\caption{Complexities for Centralized Convex Problems. This table compares communication and sample complexities in deterministic and stochastic settings. \textcolor{red}{$^\star$}Only minibatch Accelerated SGD uses minibatches; all other algorithms use local steps. NA in the ``Linear Speedup'' column indicates that the stochastic setting is not considered.}

		\renewcommand{\arraystretch}{1.49}
  \hspace*{-1em}
		\begin{threeparttable}[b]
			\begin{tabular}{c | c | c | c | c | c}
				\hline
				\hline
                     \textbf{Algorithms} &
				\textbf{Deter. Comm.} & 
				\textbf{Deter. Samp.}  & 
				\textbf{Stoc. Comm.}  &
                     \textbf{Stoc. Samp.} &
                    \textbf{Linear speedup} 
				\\
				\hline 
				\hline
				\hspace{-1mm} Minibatch Acc-SGD~\citep{woodworth2020minibatch}$\textcolor{red}{^\star}$
				& $ \cO(\epsilon^{-\frac{1}{2}})$
				& $ \cO(\epsilon^{-\frac{1}{2}})$
				&  $ \cO(\epsilon^{-\frac{1}{2}})$
                    & $ \cO(\epsilon^{-2})$ 
                    & Yes
                    \\
				\hline
		      SCAFFOLD~\citep{karimireddy2020scaffold}
				& $ \cO(\epsilon^{-1})$
				& $ \cO(\epsilon^{-1})$
				&  $ \cO(\epsilon^{-1})$
                    & $ \cO(\epsilon^{-2})$ 
                    & Yes
                    \\
				\hline
		      RandProx~\citep{condat2022randprox}
				& $ \cO(\epsilon^{-1})$
				& $ \cO(\epsilon^{-1})$
				& N/A
                    & N/A
                    & N/A
                    \\
				\hline
                   \textcolor{blue}{Catalyst+GA-MSGD}
				& \textcolor{blue}{$\widetilde \cO(\epsilon^{-\frac{1}{2}})$}
				& \textcolor{blue}{$\widetilde \cO(\epsilon^{-\frac{1}{2}})$}
				&  \textcolor{blue}{$\widetilde \cO(\epsilon^{-\frac{1}{2}})$}
                    & \textcolor{blue}{$\widetilde \cO(\epsilon^{-\frac{5}{2}-\alpha}), \alpha>0$}
                    & \textcolor{blue}{No}
                     \\
				\hline
                   \hspace{-2mm}\textcolor{blue}{Catalyst+SCAFFOLD}
				& \textcolor{blue}{$\widetilde \cO(\epsilon^{-\frac{1}{2}})$}
				& \textcolor{blue}{$\widetilde \cO(\epsilon^{-\frac{1}{2}})$}
				&  \textcolor{blue}{$\widetilde \cO(\epsilon^{-\frac{1}{2}})$}
                    & \textcolor{blue}{$\widetilde \cO(\epsilon^{-\frac{5}{2}-\alpha}), \alpha>0$} 
                    & \textcolor{blue}{Yes}
                    \\
				\hline
				Lower bound \citep{woodworth2020minibatch}
				& $\Omega(\epsilon^{-\frac{1}{2}})$
				& $\Omega(\epsilon^{-\frac{1}{2}})$
				&  $\Omega (\epsilon^{-\frac{1}{2}})$
                    & $\Omega(\epsilon^{-2})$ 
                    & Yes
                    \\
				\hline
				\hline
			\end{tabular}
		\end{threeparttable}

  \label{table:summary_results_convex}
 \end{table*}

\paragraph{Our Contributions.} We directly apply a primal-dual algorithm to the Lagrangian reformulation of problem (\ref{objective}). It runs multiple Stochastic Gradient Descent (MSGD) iterations on the primal variable as the inner loop, followed by a single (accelerated) Gradient Ascent (GA) step on the dual variable—this method is referred to as (Acc-)GA-MSGD. Since the computation of the primal gradient does not require communication between clients, this method naturally incorporates local steps.
\vspace{1mm}

(A) Centralized: When the objective is strongly convex, we show that GA-MSGD achieves linear convergence in the number of outer loops by establishing that the dual function of the Lagrangian is strongly concave. This is attributed to the coupling matrix between the primal and dual variables being full-rank. When integrated it with the Catalyst \citep{lin2018catalyst}, notable results include: 
(a) the first local method that attains \(\widetilde\cO(\sqrt \kappa)\) communication and gradient complexities simultaneously for deterministic strongly convex problems; (b) the first \textit{non-asymptotic} $\widetilde\cO(\sqrt{\kappa})$ communication complexity without minibatches for stochastic strongly convex problems. As a side result, we combine Catalyst with SCAFFOLD \citep{karimireddy2020scaffold} to further attain linear speed-up\footnote{In this work, we define an algorithm as having linear speedup if its sample complexity includes all terms associated with stochastic gradient variance, denoted as 
$\sigma^2$, divided by the total number of clients.}. Results for strongly convex and convex settings are summarized in Table \ref{table:summary_sc} and \ref{table:summary_results_convex}, respectively. 

\vspace{1mm}
(B) Decentralized:  When the objective is strongly convex, we demonstrate that Acc-GA-MSGD achieves linear convergence by observing that the dual function of the Lagrangian is strongly concave within the span of the coupling matrix between the primal and dual variables. By integrating Acc-GA-MSGD with the Catalyst framework, we achieve near-optimal communication efficiencies across strongly convex, convex, and nonconvex settings, provided that sufficient local steps are taken. This matches the lower bounds established in \citep{scaman2017optimal, scaman2019optimal, lu2021optimal}. 
The results are summarized in Table~\ref{table:summary_results1}.

\vspace{1mm}
In summary, our results reveal that by employing suitable local methods, we can achieve the near-optimal communication complexities in stochastic settings, \textit{without the need for large minibatches}. Furthermore, a straightforward minimax optimization algorithm applied to the Lagrangian can serve as an effective local method for achieving such outcomes. While the algorithms resemble the (accelerated) dual ascent methods in \citep{scaman2017optimal, terelius2011decentralized}, direct access to the gradient of the conjugate function is not available in our work.

\begin{table*}[t]
		\centering
		\small
\caption{Communication Complexities for Decentralized Stochastic Problems. This table compares communication complexity when algorithms can use sufficient local samples for minibatches or local steps. $\textcolor{red}{^\star}$assumes a heterogeneity measure is bounded by $\varsigma$. $\textcolor{red}{^\Delta}$minibatch is needed for the best communication complexity. $\kappa$ represents the condition number, and $p \triangleq 1-\sigma_2$, where $\sigma_2$ is the second largest singular value of the network weight matrix $W$. $M$ is the number of clients. $C_0$ in D-MSAG is specified in \citep{fallah2022robust}. Stochastic GT \citep{koloskova2021improved} has another network-related constant $c$ with $c \geq 1-\sigma_2$. 
  }
		\renewcommand{\arraystretch}{1.49}
  \hspace*{-1em}
		\begin{threeparttable}[b]
			\begin{tabular}{c | c | c | c }
				\hline
				\hline
				\textbf{Algorithms} & 
				\textbf{Strongly Convex}  & 
				\textbf{Convex}
				&
                    \textbf{Nonconvex}  
				\\
				\hline 
				\hline
				Local-DSGD~\citep{koloskova2020unified}$\textcolor{red}{^\star}$
				& $\widetilde \cO(\kappa p^{-1}+\sqrt{\kappa}\varsigma p^{-1}\epsilon^{-\frac{1}{2}})$
				& $\cO(p^{-1}\epsilon^{-1} + \varsigma p^{-1}\epsilon^{-\frac{3}{2}})$ 
				&  $\cO(p^{-1}\epsilon^{-2} + \varsigma p^{-1}\epsilon^{-3})$
                    \\
				\hline
                D-PSGD~\citep{lian2017can}$\textcolor{red}{^\star}$$\textcolor{red}{^\Delta}$
				& N/A
				& N/A
				&  $ \cO(\varsigma M p^{-2}c^{-1}\epsilon^{-2})$
                    \\
				\hline
                DeTAG~\citep{lu2021optimal}$\textcolor{red}{^\star}$$\textcolor{red}{^\Delta}$
				& N/A
				& N/A
				&  $ \widetilde\cO(p^{-\frac{1}{2}}\epsilon^{-2})$
                    \\
				\hline
                D-MSAG~\citep{fallah2022robust}$\textcolor{red}{^\Delta}$
				& $\widetilde \cO(\kappa^{\frac{1}{2}}p^{-\frac{1}{2}} + C_0\sqrt{\kappa}p^{-\frac{1}{2}}(M\epsilon)^{-\frac{1}{4}})$
				& N/A
				&  N/A
                    \\
				\hline
				Stochastic GT~\citep{koloskova2021improved}$\textcolor{red}{^\Delta}$
				& $\widetilde \cO( \kappa p^{-1}c^{-1})$
				& $ \cO(   p^{-1}c^{-1}\epsilon^{-1})$
				&  $ \cO(   p^{-1}c^{-1}\epsilon^{-2})$
                    \\
				\hline
				K-GT~\citep{liu2024decentralized}
				& N/A
				& N/A
				&  $ \cO(  p^{-2}\epsilon^{-2})$
                    \\
				\hline				LED~\citep{alghunaim2024local}
				& $\widetilde \cO(\kappa^2 p^{-1})$
				& $\cO( p^{-1}\epsilon^{-1}))$
				&  $\cO( p^{-1}\epsilon^{-2})$
                    \\
				\hline
				\textcolor{blue}{Acc-GA-MSGD}
				& \textcolor{blue}{$\widetilde \cO(\kappa^{\frac{1}{2}} p^{-\frac{1}{2}})$}
				& \textcolor{blue}{N/A}
				&  \textcolor{blue}{N/A}
                    \\
				\hline
				\hspace{-2mm} \textcolor{blue}{Catalyst+Acc-GA-MSGD}
				& \textcolor{blue}{$\widetilde \cO(\kappa^{\frac{1}{2}} p^{-\frac{1}{2}})$}
				& \textcolor{blue}{$\widetilde\cO( p^{-\frac{1}{2}}\epsilon^{-\frac{1}{2}})$}
				&  \textcolor{blue}{$\widetilde\cO( p^{-\frac{1}{2}}\epsilon^{-2})$}
                    \\
				\hline
				Lower bound \citep{scaman2017optimal, scaman2019optimal, lu2021optimal}
				& $\widetilde\Omega(\kappa^{\frac{1}{2}} p^{-\frac{1}{2}})$
				& $\Omega( p^{-\frac{1}{2}}\epsilon^{-\frac{1}{2}})$
				&  $\Omega( p^{-\frac{1}{2}}\epsilon^{-2})$
                    \\
				\hline
				\hline
			\end{tabular}
		\end{threeparttable}

  \label{table:summary_results1}
 \end{table*}

\subsection{Related Work}

\paragraph{Strongly convex:} 
In the centralized setup, local methods such as ProxSkip (also known as Scaffnew)~\citep{mishchenko2022proxskip} and DualFL~\citep{park2023dualfl} attain optimal communication complexity \(\widetilde\cO(\sqrt{\kappa})\) but suboptimal gradient complexity \(\widetilde \cO(\kappa)\) in the deterministic setting~\citep{mishchenko2022proxskip}. In the stochastic setting, the communication complexity of ProxSkip was recently tightened to the near-optimal \(\widetilde\cO(\sqrt{\kappa})\), though only asymptotically~\citep{hu2023tighter}. 
In the decentralized setup, deterministic algorithms have achieved near-optimal communication complexity \(\widetilde\Omega(\kappa^{\frac{1}{2}}(1-\sigma_2)^{-\frac{1}{2}})\)~\citep{scaman2017optimal, uribe2020dual, li2020decentralized, li2020revisiting, song2023optimal}. Notably, \citet{uribe2020dual} incorporated local steps in an algorithm similar to Scaffnew. \citet{fallah2022robust} considered decentralized multi-stage Accelerated Stochastic Gradient (D-MASG) that can converge to the exact solution in the stochastic setting. The stochastic variant of Gradient Tracking~\citep{pu2018distributed} with sufficiently large minibatches achieves a communication complexity of \(\widetilde\cO(\frac{\kappa}{(1-\sigma_2)c})\), where \(c \geq \Theta(1-\sigma_2)\) is a network-related constant~\citep{koloskova2021improved}. 
Recently, Exact Diffusion with local steps achieved a complexity of \(\widetilde\cO(\frac{\kappa^2}{1-\sigma_2})\)~\citep{alghunaim2024local}.

\vspace{-2mm}
\paragraph{Convex:} 
In the centralized setup, while minibatch Accelerated SGD achieves the optimal communication complexity of \(\mathcal{O}(\epsilon^{-1/2})\)~\citep{woodworth2020minibatch}, many federated learning algorithms that do not require minibatches, including SCAFFOLD~\citep{karimireddy2020scaffold} and RandProx~\citep{condat2022randprox}, remain at \(\mathcal{O}(\epsilon^{-1})\). Additionally, decentralized algorithms can attain the optimal communication complexity of \(\mathcal{O}((1-\sigma_2)^{-\frac{1}{2}} \epsilon^{-\frac{1}{2}})\) in deterministic settings~\citep{li2020decentralized, uribe2020dual, scaman2019optimal, li2020revisiting}, with some exploring algorithms with local steps~\citep{uribe2020dual}. A stochastic algorithm with local steps achieves a communication complexity of \(\mathcal{O}((1-\sigma_2)^{-1} \epsilon^{-1})\)~\citep{alghunaim2024local}. 

\vspace{-2mm}
\paragraph{Nonconvex:} 
In the centralized setting, under bounded heterogeneity assumptions, local SGD attains a gradient complexity of \(\cO(\epsilon^{-4})\) and a suboptimal communication complexity of \(\cO(\epsilon^{-3})\) \citep{yu2019linear}. Several algorithms, including \citep{karimireddy2020scaffold, cheng2023momentum, zhang2021fedpd, alghunaim2024local, liang2019variance}, achieve near-optimal communication and gradient complexities without bounded heterogeneity. Some of them additionally achieve linear speedup \citep{cheng2023momentum}. In the decentralized setting, local SGD suffers from suboptimal communication complexity in $\epsilon$~\citep{koloskova2020unified}, while decentralized minibatch SGD achieves \(\cO((1-\sigma_2)^{-2}\epsilon^{-2})\) with bounded heterogeneity~\citep{lian2017can}. It was further improved with techniques such as \textsc{PushSum}~\citep{assran2019stochastic} and gradient tracking~\citep{xin2021improved}. The lower bound of $\Omega((1-\sigma_2)^{-\frac{1}{2}}\epsilon^{-2}))$ communication complexity was established in \citep{lu2021optimal}, and a nearly matched algorithm was also provided, but requires a large batch size.  Gradient Tracking and Exact Diffusion, when combined with local steps, achieve communication complexities of \(\cO((1-\sigma_2)^{-2}\epsilon^{-2})\) \citep{liu2024decentralized} and \(\cO((1-\sigma_2)^{-1}\epsilon^{-2})\) \citep{alghunaim2024local}, respectively.

\vspace{-2mm}
\paragraph{(Primal)-Dual Algorithms:}
There is a long history of using primal-dual algorithms to solve the (augmented) Lagrangian. The Alternating Direction Method of Multipliers (ADMM) is a popular technique for solving the augmented Lagrangian~\citep{boyd2011distributed, wei20131, makhdoumi2017convergence}. Several works have suggested using linearized primal subproblems~\citep{chang2014multi, hong2017stochastic, ling2015dlm, aybat2017distributed}, with a closed form for the primal update that involves a one-step (stochastic) gradient descent in the primal. Many algorithms, including EXTRA~\citep{shi2015extra, mokhtari2016dsa, li2020revisiting} and DIGing~\citep{nedic2017achieving}, are interpreted or motivated by the Gradient Descent Ascent algorithm on the augmented Lagrangian. The quadratic penalty term related to the gossip matrix in the augmented Lagrangian prevents a straightforward extension of these methods to include multiple gradient descent steps in the primal variable without communication. These methods also closely resemble linearized ADMM~\citep{hong2017stochastic}. \citet{mokhtari2016dsa} showed linear convergence for stochastic strongly convex problems when each agent has finite-sum objectives by incorporating incremental averaging gradients used in variance-reduction literature. \citet{lan2020communication} considers a primal-dual method on the non-augmented Lagrangian for non-smooth convex problems. The most closely related works to ours are \citep{terelius2011decentralized, ghadimi2011accelerated, scaman2017optimal, uribe2020dual}, which executed (accelerated) dual ascent in the dual problem. There are two main differences between our work with the first three works: they assume the exact dual gradients are available, whereas we use stochastic gradient descent in the primal function to find inexact dual gradients; and their convergence is explicitly provided only for the dual variable, rather than the primal variable. Specifically, linear convergence is provided in \citep{scaman2017optimal}, but the proof uses the Hessian of the conjugate function, whose existence requires additional assumptions, such as twice differentiability of the original objective. \citet{uribe2020dual} used Accelerated Gradient Descent to approximate the dual gradient but did not address stochastic or non-convex settings.


\section{Preliminary}

\begin{algorithm}[t] 
    \caption{Catalyst \citep{lin2018catalyst, paquette2018catalyst}}
    \begin{algorithmic}[1] 
        \State \textbf{Initialization}: initial point $x_0$. In the strongly convex setting, $q = \frac{\mu}{\mu + 2L}$ and $\alpha_1 = \sqrt{q}$; in the convex and nonconvex settings, $q = 0$ and $\alpha_1 = 1$. Initialize $y^0 = x^0$.
        \For{$s = 1,2,..., S$}
            \State Find an inexact solution $x^s$ to the following problem with an algorithm $\mathcal{A}$
            \vspace{-1mm}
            \begin{equation} \label{cata:subproblem}
            \min_{x} \left[F^s(x) \triangleq F(x) + L\|x - y^s\|^2 \right].
            \vspace{-1mm}
            \end{equation} 
           \quad \ \ such that $\bE F^s(x^s) - \min_x F^s(x) \leq \epsilon^s$. \vspace{1mm}
            \State Update momentum parameters: in the strongly convex and convex settings, $\alpha_s^2=\left(1-\alpha_s\right) \alpha_{s-1}^2+q \alpha_s$ and $\beta_s=\frac{\alpha_{s-1}\left(1-\alpha_{s-1}\right)}{\alpha_{s-1}^2+\alpha_s}$; in the nonconvex setting, $\beta_s = 0$.
            \State Compute the new prox center: $y^{s+1} = x^s + \beta_s(x^s - x^{s-1})$
        \EndFor
    \end{algorithmic} 
\label{alg:APPA}
\end{algorithm}

\paragraph{Notation and Acronym} By default, we use the Frobenius norm and  inner product 
$\langle A, B\rangle = Tr(AB^\top)$ for matrices, and the Euclidean norm for vectors. The symbol $\one$ denotes a column vector consisting of all ones with an appropriate dimension. We use the acronyms (SC) for strongly convex, (C) for convex, and (NC) for nonconvex.

The Catalyst framework \citep{lin2018catalyst} embodies an inexact proximal point method \citep{guler1991convergence}, as outlined in Algorithm~\ref{alg:APPA}. It is applicable to objective functions that are strongly convex, convex, or nonconvex. At each round $s$, the algorithm updates $x^s$ to be an approximate solution of a strongly convex subproblem (\ref{cata:subproblem}), which integrates a quadratic regularization term. The center of regularization, $y^s$, is selected using an extrapolation step ($\beta_s > 0$) in strong convex and convex settings, and is selected to be $x^s$ without any momentum ($\beta_s = 0$) in the nonconvex setting. In the following theorem, we review the convergence guarantees for Catalyst, which represents a slight modification of \citep{lin2018catalyst, paquette2018catalyst}. We replace the subproblem stopping criterion $F^s(x^s) - \min_x F^s(x) \leq \epsilon^s$ in \citep{lin2018catalyst} by its counterpart in expectation, and the same results still hold but in expectation.

\begin{theorem}
\label{thm:catalyst}
    Assume that $F$ is $L$-smooth. Define $F^* = \min_x F(x)$ and $\Delta = F(x^0) - F^*$. The following convergence guarantees for Algorithm \ref{alg:APPA} hold with the specified subproblem accuracy $\{ \epsilon^s\}_s$ under different convexity assumptions for $F$.
    \begin{itemize}[leftmargin=*, itemsep=0pt]
        \item \textbf{(SC)}:  With $\epsilon^s = \frac{2(1-0.9\sqrt{q})^s\Delta}{9}$ and $q  = \frac{\mu}{\mu + 2L}$, then $\bE F\left(x^S\right)-F^* \leq \frac{800(1-0.9\sqrt{q})^{S+1}\Delta}{q} $.
        \item \textbf{(C)}: With $\varepsilon^s=\frac{2\Delta}{9(s+1)^{4+\gamma}}$ and $\gamma > 0$, $
            \bE F\left(x^S\right)-F^* \leq \frac{8}{(S+1)^2}\left(L\left\|x^0-x^*\right\|^2+\frac{4\Delta}{\gamma^2}\right)$.
        \item \textbf{(NC)}: with $\epsilon^s = \frac{\Delta}{S}$, then
        $
            \frac{1}{S}\sum_{s=1}^{S} \bE \|\nabla F(x^s)\|^2 \leq \frac{32L\Delta}{S}.
        $
    \end{itemize}
\end{theorem}

\citet{li2020revisiting} extended the Catalyst framework to decentralized optimization by demonstrating that applying Catalyst individually to each client essentially equates to applying the original Catalyst to the average of the clients' variables (see Section \ref{subsec:cata-decentral} for more details). This adaptation allows us to apply Theorem~\ref{thm:catalyst} in decentralized settings as well.

In this paper, when we consider the distributed problem of form (\ref{objective}), we adopt the following umbrella assumptions, which is common in the literature. 

\begin{assumption}
\label{assum:global}
(1) Local function of each client is $L$-Lipsthiz smooth, i.e., $\|F_m(x) - F_m(\hat x)\| \leq L \|x - \hat x\|$ for all $x$ and $\hat x$. (2) Each client $m$ has access to the unbiased stochastic gradient $g_m(x; \xi)$ of its local function such that
    $\bE_{\xi} \ g_m(x; \xi) = \nabla F_m(x)$ and $
    \bE_{\xi} \ \|g_m(x; \xi) - \nabla F_m(x)\|^2 \leq \sigma^2.
$
\end{assumption}

\vspace{-2mm}
\section{Centralized Optimization}

In the centralized setup, there exists a central coordinator that can communicate with all clients. Without loss of generalization, we assume the first client is the central coordinator. This communication includes broadcasting information to the clients and aggregating information received from them. We consider a constrained optimization reformulation where each client maintains its local decision variable $x_m$. These variables are bound by a consensus constraint ensuring that all local variables are equal to the global variable held by the coordinator:
\vspace{-2mm}
\begin{align*}
    \min _{x_1, x_2, \dots, x_M \in \bR^n}  \ \frac{1}{M} \sum_{m=1}^M F_m(x_m), \quad \text{subject to } \ x_m = x_1, \text{for } m = 2,3,\dots, M.
    \vspace{-1mm}
\end{align*}
For the ease of the presentation, we take the following notations: 
$$
H(x_1, X)= \frac{1}{M}\sum_{i=1}^M F_m\left(x_m\right), \quad X=\left(\begin{array}{c}
x_{2}^\top \\
\vdots \\
x_{M}^\top
\end{array}\right)\in \bR^{(M-1)\times n}, \quad \nabla H(x_1, X)=\left(\begin{array}{c}
\frac{1}{M}\nabla F_1\left(x_{1}\right)^\top \\
\vdots \\
\frac{1}{M} \nabla F_M\left(x_{M}\right)^\top
\end{array}\right)\in \bR^{M\times n} .
$$

\subsection{A Primal-Dual Algorithm}
\label{subsec:PD-alg}

\begin{assumption}
\label{assum:sc}
In this subsection, we assume the local function of each client is $\mu$-strongly convex, i.e., $F_m(\hat x) \geq F_m(x) + \langle \nabla F_m(x), \hat x - x\rangle + \frac{\mu}{2}\|\hat x - x\|^2$ for all $x$ and $\hat x$. 
\end{assumption}
By introducing the dual variable $\lambda = (\lambda_2 \ \lambda_3  \dots \lambda_M)^\top \in \bR^{(M-1)\times n}$ with each $\lambda_m \in \bR^n$ corresponding to the constraint $x_m = x_1$, the Lagrangian of the constrained formulation is:
\begin{equation*}
    \min_{x_1, X} \max_{\lambda} \  \cL(x_1, X, \lambda)
     =  H(x_1, X) + \langle
     \lambda, X - \one x_1^\top \rangle.
\end{equation*}
We further reformulate the Lagrangian by subtracting a quadratic term from each client's variable and adding a quadratic term to the coordinator's variable $x_1$. When the consensus constraint is met (i.e., $x_1 = x_2=\cdots = x_M$), the primal solution will be identical to the previous one. We opted for this reformulation because it improves the conditioning of our dual function by making the Lagrangian more strongly convex in $x_1$ (see Lemma~\ref{lemma:dual-function}).
\vspace{-2mm}
\begin{align}  
\label{eq:lagrangian_cen}
    \min_{x_1, X} \max_{\lambda} \ \widetilde \cL(x_1, X, \lambda)
     =  H(x_1, X) - \frac{\mu}{4M}\|X\|^2 + \frac{\mu(M-1)}{4M}\|x_1\|^2 + 
     \langle
     \lambda, X - \one x_1^\top \rangle.
\end{align}
The gradient of this function can be computed as follows:
\begin{align*} 
\setlength{\jot}{-1pt}
  & \nabla_{x_1}  \widetilde\cL(x_1, X, \lambda) = \frac{1}{M} \nabla F(x_1) + \frac{\mu(M-1)}{2M}x_1 - \sum_{m=2}^M \lambda_m\\ 
    &  \nabla_{x_m} \widetilde\cL(x_1, X, \lambda) = \frac{1}{M}\nabla F_m(x_m)  - \frac{\mu}{2M}x_m+  \lambda_m , \quad m = 2,3,\dots, M \\ 
    &\nabla_{\lambda_m}\widetilde \cL(x_1, X, \lambda) = x_m - x_1, \quad m = 2,3,\dots, M.
\end{align*}

We observe that once fixing $\lambda$, the gradient in $x_m$ can be computed locally. That is, worker $m$ only needs the (stochastic) gradient of its local function $F_m$. However, the gradient in the dual variable requires communications to compute the difference between global variable $x_1$ and local variables. This motivates us to consider a basic minimax optimization algorithm named Gradient Ascent Multi-Stochastic Gradient Descent (GA-MSGD), presented in Algorithm \ref{alg:multi-agda}.

This algorithm consists of two loops. In the inner loop, it performs multiple steps of Stochastic Gradient Descent (SGD) on primal variables, each of which can be executed locally by each client. Subsequently, the coordinator broadcasts its variable, $x_1$, to all clients, who then update their dual variable, $\lambda_m$, by performing a single step of Gradient Ascent. All clients send their $\lambda_m$ back to the coordinator to start the next iteration.

While this algorithm is described using Stochastic Gradient Descent (SGD) to solve for the primal variables, formula (\ref{eq:lagrangian_cen}) allows for considerable flexibility in selecting optimization methods. Alternatives to SGD, such as adaptive methods like Adam~\citep{KingBa15} or Shampoo~\citep{gupta2018shampoo}, can be employed. Furthermore, it is possible for each client to use different optimizers in local steps.

\begin{algorithm}[t]  
    \caption{Gradient Ascent Multi-Stochastic Gradient Descent (GA-MSGD)}
    \begin{algorithmic}[1] 
      \State \textbf{Input:} initial point $x^0$
      and $\{\lambda_m^0\}_{m=1}^M$.
      \State \textbf{Initialize} $x_m^{0,0} = x^0$ for $m=1, 2,\dots, M$
        \For{$t = 0,1,2,\dots, T$}
            \For{$k = 0,1,2,..., K-1$}
                 \State sample  $\{\xi^{k}_m\}_{m=1}^M$
                  \State  $x_m^{t, k+1} = x_m^{t, k} -\tau_1^k\left[\frac{1}{M}g_m(x_m^{t, k}; \xi_m^k) - \frac{\mu}{2M}x_m^{t,k} + \lambda_m^t  \right]$ for $m=2,\dots, M$
                  \State $x_1^{t, k+1} = x_1^{t, k} - \tau_2^k\left[\frac{1}{M} g_1(x_1^{t, k};\xi_1^k) + \frac{\mu(M-1)}{2M}x_1^{t,k}  - \sum_{m=2}^M \lambda_m^t \right]$
            \EndFor
            \State $x_m^{t+1, 0} = x_m^{t, K}$ for $m=2,3,\dots, M$    
            \State $\lambda_m^{t+1} = \lambda_m^t + \tau_3 \left[x_m^{t+1, 0} - x_1^{t+1, 0}\right]$ for $m=2,3,\dots, M$   
        \EndFor
        \State \textbf{Output:} $(\bar x^T, \lambda^T)$ with $\bar x^T = \frac{1}{M}\sum_{m=1}^M x_m^{T, K}$  
    \end{algorithmic} 
\label{alg:multi-agda}
\end{algorithm}



\subsection{Convergence Analysis for Centralized GA-MSGD}

In Algorithm~\ref{alg:multi-agda}, communications occur only within the outer loop. We now present the convergence of the outer loop, which will characterize the communication complexity of GA-MSGD.

\begin{theorem}[Outer-loop Complexity]
\label{thm:outer-loop-central}
Under Assumption~\ref{assum:global} and \ref{assum:sc}, we choose inner loop iterations $K$ large enough such that $\bE \sum_{m=2}^M \|x_m^{t, K} - x_m^*(\lambda^t)\|^2 \leq \delta_1$ and $\bE \|x_1^{t, K} - x_1^*(\lambda^t)\|^2 \leq \delta_2$, where $(x_1^*(\lambda^t), x_2^*(\lambda^t), \dots, x_M^*(\lambda^t)) = \argmin_{x_1, X} \widetilde\cL(x_1, X, \lambda^t)$ is the optimal primal variable corresponding to $\lambda^t$. If we initialize $\lambda_m^0 = - \frac{1}{M}g_m(x^0;\xi)  + \frac{L}{2M}x^0$ for all $m$, the output $\bar x^T$ satisfies
\begin{equation*}
     \bE F(\bar x^T) - F^* \leq  \frac{48L^4}{\mu^3}  \left(1 - \frac{\mu}{6L} \right)^T  \left[\|x^0-x^*\|^2 + \frac{2\sigma^2M}{L^2} \right]   + \frac{40L^3\delta_1}{\mu^2 M}  + \frac{40L^3\delta_2}{\mu^2 },
\end{equation*}
where $x^*$ is the optimal solution to the original problem (\ref{objective}).
\end{theorem}

The theorem implies that GA-MSGD converges linearly, up to an error term due to the inexactness of the inner loops.  
Note that the Lagrangian is linear in $\lambda$, and the optimal complexity of solving a general minimax optimization problem that is only linear in the dual variable is $\cO(\epsilon^{-\frac{1}{2}})$ \citep{ouyang2021lower}. Here linear convergence arises because we can show the dual function $\widetilde\Psi(\lambda) \triangleq \min_{x_1, X} \widetilde\cL(x_1, X, \lambda)$ is strongly concave by leveraging the following proposition.  A similar argument is presented in \citep{du2019linear, ghadimi2011accelerated}.

\begin{prop}
\label{prop:duality}
Assume the function \( f: \mathbb{R}^{m \times n} \rightarrow \mathbb{R} \) is \( \nu \)-strongly convex and \( \ell \)-smooth. Let \( A \in \mathbb{R}^{d\times m} \) be a matrix, with \( \sigma_{\max} \) and \( \sigma_{\min} \) representing the maximum and minimum singular values of \( A \), respectively. Then, the function \( h(\lambda) = \min_x \{f(x) + \langle \lambda, Ax \rangle\} \) is \( \sigma_{\max}^2/\nu \)-smooth and concave. When \( A \) is full-row rank, \( h \) is further \( \sigma_{\min}^2/\ell \)-strongly concave. Additionally,  \( x^*(\lambda) \triangleq \arg\min_x \{f(x) + \langle \lambda, Ax \rangle\} \) is \( \sigma_{\max}/\nu \)-Lipschitz in \( \lambda \).
\end{prop}

Observing that in (\ref{eq:lagrangian_cen}) the coupling matrices between $\lambda$ with primal variables, $X$ and $x_1$, are identity matrix and $\one$, respectively, we use this proposition to establish that the dual function $\widetilde \Psi(\lambda)$ is $\frac{4M}{\mu}$-smooth and $\frac{2M}{3L}$-strongly concave in Lemma~\ref{lemma:dual-function}. \citet{du2019linear} demonstrated linear convergence using deterministic Gradient Descent Ascent (GDA) for the class of minimax problems with a full-rank coupling matrix. We opted for GA-MSGD since we only have access to stochastic gradients for the primal variables. By integrating an inner loop of Stochastic Gradient Descent, we can execute local updating steps for each client, while preserving the linear convergence of the outer loop. Furthermore, the function $\widetilde\cL$ is separable in $X$ and $x_1$, which aids in achieving a better condition number for linear convergence.

Theorem \ref{thm:outer-loop-central} shows that the algorithm will still converge even if optimizers other than SGD are used in the inner loop, as long as the required accuracy for these loops is achieved. For simplicity, we present the inner-loop complexity when every client uses SGD as local steps, following the SGD convergence for strongly convex function~\citep{stich2019unified}.

\begin{lemma}[Inner-loop Complexity]
\label{thm:inner-loop-central}
Choosing $\delta_1 = \frac{\mu^2 M \epsilon}{120L^3}$ and $\delta_2 = \frac{\mu^2 \epsilon}{120L^3}$ in Theorem~\ref{thm:outer-loop-central}, with proper stepsize $\{\tau_1^k\}_k$ and $\{\tau_2^k\}_k$, the number of inner loop iterations  (where $\kappa = L/\mu$ is the condition number)
\begin{equation*}
     K = \cO\left(\kappa\log\left(\frac{M\kappa \sigma^2}{\mu \epsilon} + \frac{\kappa L\|x^0 - x^*\|^2}{\epsilon} + \kappa \right) + \frac{\kappa^3\sigma^2}{\mu\epsilon} \right).
\end{equation*}
\end{lemma}

\begin{corollary} [Total Complexity]
\label{coro:total-complexity}
   To find an $\epsilon$-optimal point $x$ such that $\bE \|F(x) - F^* \| \leq \epsilon$, the total number of communication rounds in Algorithm \ref{alg:multi-agda} is 
   \begin{equation*}
       T = \cO \left( \kappa \log\left( \frac{\kappa L\|x^0 - x^*\|^2}{\epsilon} + \frac{M\sigma^2}{\mu\epsilon} \right)\right).
   \end{equation*}
   The total number of local gradients (samples) used by each client is 
   \begin{equation*}
       TK = \cO \left( \kappa \log^2\left(\frac{M\kappa \sigma^2}{\mu \epsilon} + \frac{\kappa L\|x^0 - x^*\|^2}{\epsilon} + \frac{\kappa}{\epsilon}  \right) + \frac{\kappa^4\sigma^2}{\mu\epsilon}  \log\left( \frac{\kappa L\|x^0 - x^*\|^2}{\epsilon} + \frac{M\sigma^2}{\mu\epsilon} \right)\right).
   \end{equation*}
\end{corollary}

The corollary presents a communication complexity of \(\widetilde\cO(\kappa)\). Given that we establish the condition number for the dual function \(\widetilde \Psi(\lambda)\) as \(6\kappa\) in Lemma~\ref{lemma:dual-function}, applying Accelerated Gradient Ascent \citep{nesterov2013introductory} in the dual update of Algorithm~\ref{alg:multi-agda} will lead to an improved communication complexity of \(\widetilde\cO(\sqrt{\kappa})\). However, as we intend to refine the \(\kappa\) dependency with Catalyst in the subsequent subsection, we continue with Gradient Ascent here.

\subsection{Catalyst Acceleration for GA-MSGD}

We integrate GA-MSGD with Catalyst by employing it as algorithm $\mathcal{A}$ to solve the strongly convex subproblem (\ref{cata:subproblem}). It will not only enhances the dependency on the condition number in the strongly convex setting but also empowers GA-MSGD to solve convex and nonconvex problems. This framework can be easily implemented in a centralized way: after the coordinator receives the solution $x^s$, it computes the new proximal center $y^{s+1}$ and broadcast it. Then all clients solve the new subproblem with local function $F_m^s(x) = F_m(x) + L\|x - y^{s+1}\|^2$.

\begin{corollary}
\label{coro:catalyst-GA-MSGD}
    If we solve the subproblem (\ref{cata:subproblem}) in Catalyst (Algorithm~\ref{alg:APPA}) with Algorithm~\ref{alg:multi-agda}, the communication complexity and sample complexities are as follows: where (SC), (C), and (NC) denote that each client's local function is strongly convex, convex, and nonconvex, respectively,
    \begin{itemize}[leftmargin=*, itemsep=0pt]
\item \textbf{(SC) } To achieve $\bE F(x) - F^*\leq \epsilon$: \ comm: $\widetilde\cO(\sqrt{\kappa})$; \ sample: $\widetilde\cO(\sqrt{\kappa} + \frac{\sqrt\kappa\sigma^2}{\mu\epsilon})$.
\item \textbf{(C) } To achieve $\bE F(x) - F^*\leq \epsilon$:  \ comm: $\widetilde\cO(\frac{1}{\sqrt{\epsilon}})$; \ sample: $\widetilde\cO(\frac{1}{\sqrt{\epsilon}} + \frac{\sigma^2}{\epsilon^{2.5+\gamma/2}})$ for $\gamma \in (0,1]$.
\item \textbf{(NC) }  To achieve $\bE \|\nabla F(x)\|^2\leq \epsilon^2$: \ comm: $\widetilde\cO(\frac{1}{\epsilon^{2}})$; \ sample: $\widetilde\cO(\frac{1}{\epsilon^{2}} + \frac{\sigma^2}{\epsilon^{4}})$.
\end{itemize}
\end{corollary}

\begin{remark}
\label{remark:central}
For strongly convex objectives, it achieves the optimal communication complexity $\widetilde \cO(\sqrt \kappa)$ \citep{woodworth2020minibatch, woodworth2021min} up to logarithmic terms for both deterministic ($\sigma = 0$) and stochastic settings. The existing local method can only achieve such communication complexity asymptotically in the stochastic setting \citep{hu2023tighter}. Moreover, It improves the previous best-known gradient complexity of $\widetilde \cO(\kappa^{\frac{3}{4}})$ for local methods \citep{sadiev2022communication} to near-optimal $\widetilde \cO(\sqrt \kappa)$ in the deterministic setting, and attains a sample complexity nearly optimal in $\epsilon$ for the stochastic setting. For convex objectives, it also achieves near-optimal $\widetilde\cO(\epsilon^{-\frac{1}{2}})$ communication complexity~\citep{woodworth2020minibatch}. To our knowledge, there are no existing local methods providing such complexities. For nonconvex objectives, it attains comparable complexity in the dependency of $\epsilon$ with state-of-the-art approaches such as \citep{karimireddy2020scaffold, cheng2023momentum}.
    
\end{remark}

\begin{remark}
The results above do not demonstrate linear speedup, as the term with variance $\sigma^2$ in the sample complexity does not decrease with the number of clients $M$. Existing local methods achieving linear speedup typically incorporate an interpolation step in the update of the primal variable \citep{karimireddy2020scaffold, cheng2023momentum}.  Additionally, there is room for improvement in sample complexities in the dependency on $\kappa$ for the strongly convex setting and the dependency on $\epsilon$ for the convex setting. This improvement may hinge on designing better stopping criteria in subproblems for Catalyst.
Recently, \citet{lan2023novel} designed a new stopping criterion using SGD as a subroutine to achieve the optimal sample complexities, but it requires a tailored analysis for the subroutine. The approach could potentially be combined with our distributed algorithm in the future.

\end{remark}

\subsection{Catalyst Acceleration for SCAFFOLD}

We explore the combination of Catalyst with an existing algorithm, SCAFFOLD \citep{karimireddy2020scaffold}, to achieve linear speedup. The convergence of SCAFFOLD in the strongly convex setting is summarized below, as a direct corollary from Theorem VII and Remark 10 in~\citep{karimireddy2020scaffold}.

\begin{theorem} [\citep{karimireddy2020scaffold}]
\label{thm:scaffold}
For a $\mu$-strongly convex and $L$-smooth problem, after $T$ communication rounds with $T \geq 160L/\mu$, where each client executes $K$ local update steps per round, the output, $\bar x^T$, from SCAFFOLD satisfies (where $\tilde D^2 = 2\|x^0 - x^*\|^2 + \frac{\sigma^2}{L^2}$):
\vspace{-2mm}
\begin{equation*}
    \bE F(\bar x^T) - F(x^*) \leq \widetilde \cO \left(\frac{\sigma^2}{\mu T K M} + \mu \tilde D^2 \exp{\left(-\frac{\mu T}{162L}\right)} \right),
\end{equation*}
\end{theorem}

\begin{corollary}
\label{coro:catalyst-scaffold}
    If we solve the subproblem (\ref{cata:subproblem}) in Catalyst (Algorith~\ref{alg:APPA}) with SCAFFOLD, the communication complexity and sample complexity are as follows: 
    \begin{itemize}[leftmargin=*, itemsep=0pt]
\item \textbf{(SC) } To achieve $\bE F(x) - F^*\leq \epsilon$: \ comm: $\widetilde\cO(\sqrt{\kappa})$; \ sample: $\widetilde\cO(\sqrt{\kappa} + \frac{\sqrt \kappa\sigma^2}{\mu M\epsilon})$.
\item \textbf{(C) } To achieve $\bE F(x) - F^*\leq \epsilon$: \ comm: $\widetilde\cO(\frac{1}{\sqrt{\epsilon}})$; \ sample: $\widetilde\cO(\frac{1}{\sqrt{\epsilon}} + \frac{\sigma^2}{M\epsilon^{2.5+\gamma/2}})$ for $\gamma \in (0,1]$.
\item \textbf{(NC) }  To achieve $\bE \|\nabla F(x)\|^2\leq \epsilon^2$: \ comm.: $\widetilde\cO(\frac{1}{\epsilon^{2}})$; \ sample: $\widetilde\cO(\frac{1}{\epsilon^{2}} + \frac{\sigma^2}{M\epsilon^{4}})$.
\end{itemize}
\end{corollary}

\begin{remark}
Combining SCAFFOLD with Catalyst not only realizes the advantages in Remark~\ref{remark:central} but also has the $\sigma^2$ term in sample complexity diminishes linearly with the number of clients $M$. However, compared to GA-MSGD, SCAFFOLD requires smaller step sizes for local updates (see Section C in \citet{woodworth2020minibatch}) and offers less flexibility in choosing the local optimizer.
\end{remark}

\vspace{-3mm}
\section{Decentralized Optimizaiton}

In the decentralized setup, we address a problem of form (\ref{objective}), wherein each client communicates with its immediate neighbors in a network, devoid of a master node. The network is characterized by a weight matrix $W \in \bR^{M\times M}$ satisfying the following assumption:
\begin{assumption}
\label{assume:W}
(1) $W=W^\top, I \succeq W \succeq-I $ and $  W \mathbf{1}=\mathbf{1}$.
(2) $ \sigma_2 <1$ where $\sigma_2 $ is the second largest singular value of $W$.
\end{assumption}
Here, the set of neighbors for client $m$ is denoted as $\cN_m = \{j\neq m: W_{m, j}\neq 0 \}$. Additionally, we define the matrix $U=\sqrt{I-W} \in \mathbb{R}^{M \times M}$, following \citep{shi2015extra, li2020decentralized}.
The problem can be reformulated as a constrained optimization problem:
$$
\min_{X \in \bR^{M \times n}} H(X), \quad \text { subject to } \quad UX=0,
$$
where the following notations are adopted: 
$$
H(X)=\frac{1}{M}\sum_{m=1}^M f_m\left(x_{m}\right), \quad X=\left(\begin{array}{c}
x_{1}^\top \\
\vdots \\
x_{M}^\top
\end{array}\right) \in \bR^{M\times n}, \quad \nabla H(X)=\left(\begin{array}{c}
\frac{1}{M}\nabla F_1\left(x_{1}\right)^\top \\
\vdots \\
\frac{1}{M}\nabla F_m\left(x_{M}\right)^\top
\end{array}\right) \in \bR^{M\times n} .
$$


Without a coordinator, clients cannot achieve consensus in one communication round. Therefore, we seek for a solution $(x_1, \dots, x_M)$ where their average $\bar x$ is $\epsilon$-optimal.
We also provide a bound for consensus violation $\frac{1}{M}\sum_{m=1}^M\|x_m - \bar x\|^2 \leq \cO(poly(\epsilon))$ in our proof. But if a higher consensus is needed, an average consensus algorithm ~\citep{olfati2007consensus, xiao2004fast} or its accelerated version~\citep{liu2011accelerated} can be easily run at the end of the algorithm.  To achieve $\epsilon^\prime$ consensus, the accelerated average consensus algorithm requires only $\cO((1-\sigma_2)^{-1/2}\log \frac{1}{\epsilon^\prime})$ rounds of communications \citep{liu2011accelerated}. Since the singular value $\sigma_2$ of $W$ can be very close to 1, our complexity analysis retains the dependency on $\frac{1}{1-\sigma_2}$. 


\subsection{A Primal-Dual Algorithm}

Introducing dual variable $\lambda = (\lambda_1 \ \dots \lambda_M)^\top \in \bR^{M\times n}$, the Lagrangian of the constrained form is 
\vspace{-1mm}
\begin{equation}
\label{eq:lagrangian-decen}
\min_{X \in \bR^{M \times n}} \max_{\lambda \in \bR^{M\times n}} \cL(X, \lambda) = H(X) +  \langle \lambda, UX\rangle .
\end{equation}
This formula is widely used in the literature~\citep{li2020revisiting, lan2020communication, xu2020accelerated}. Similar to Section~\ref{subsec:PD-alg}, we implement Gradient Descent for $X$ in the inner loop, while utilizing Accelerated Gradient Ascent \citep{nesterov2013introductory} for $\lambda$. This accelerated approach is employed to improve the dependency on $1-\sigma_2$.
\vspace{-1mm}
\begin{align} \nonumber
    & X^{t, k+1} = X^{t, k} - \tau_1^k\left[\nabla H(X^{t, k}) + U\widetilde\lambda^t \right], \quad k = 1, 2, \dots, K \\  \label{eq:lambda-update}
    &\lambda^{t+1} = \widetilde\lambda^t + \tau_2 UX^{t, K}, \quad \widetilde\lambda^{t+1} = \lambda^{t+1} + \beta (\lambda^{t+1} - \lambda^t).
\end{align}
Define $\zeta = U\lambda$, and pre-multiply the update rule of $\lambda$ by $U$. Then it can be written as
\begin{align*}
    & X^{t, k+1} = X^{t, k} - \tau_1^k\left[ \nabla H(X^{t, k}) + \widetilde \zeta^t \right], \quad k = 1, 2, \dots, K\\
    &\zeta^{t+1} = \widetilde\zeta^t + \tau_2 (I-W)X^{t, K}, \quad \widetilde\zeta^{t+1} = \zeta^{t+1} + \beta (\zeta^{t+1} - \zeta^t).
\end{align*}
We present Acc-GA-MSGD in Algorithm~\ref{alg:acc-multi-agda}, where $\nabla H$ is substituted with its stochastic estimate. An important observation is that the update of $X$ in the inner loops does not require communication.
Substituting $X^{t, K}$ with $X^*(\zeta^t) \triangleq \argmin_{X} H(X) + \langle \zeta^t, X \rangle$ in the update of $\zeta^{t+1}$ transforms the algorithm into Accelerated Dual Ascent~\citep{scaman2017optimal}. When the dual gradient $X^*(\lambda)$ is not directly accessible, \citet{uribe2020dual} have employed Accelerated Gradient Descent in the inner loops to approximate it.  Following a similar concept, Algorithm~\ref{alg:acc-multi-agda} employs Stochastic Gradient Descent (SGD) in the inner loops to achieve this approximation. Since the update of $X$ is decomposable across clients, each client may also independently select their optimization method, such as ADAM~\citep{KingBa15}.

\vspace{-2mm}
\paragraph{Comprison with other primal-dual methods} Instead of directly solving the Lagrangian (\ref{eq:lagrangian-decen}), many decentralized algorithms are formulated or can be interpreted as solving the augmented Lagrangian:
\[
\min_{X \in \bR^{M \times n}} \max_{\lambda \in \bR^{M\times n}} \cL(X, \lambda) = H(X) +  \langle \lambda, AX\rangle + \alpha\|AX\|^2,
\]
where $AX$ represents the consensus constraint for some matrix $A$ related to $W$. Examples of such algorithms include 
EXTRA~\citep{hong2017prox, shi2015extra} and DIGing~\citep{nedic2017achieving}. While this quadratic term enhances the convexity condition, it usually requires communication in the update of primal variables.

\vspace{-2mm}
\paragraph{Comparison with LED and Exact Diffusion} Local Exact-Diffusion (LED), presented in~\ref{alg:led}, was recently introduced by \citet{alghunaim2024local}. It is equivalent to performing Gradient Ascent on the dual variable in the Lagrangian (\ref{eq:lagrangian-decen})—specifically, setting the momentum ($\beta$) to zero in Algorithm~\ref{alg:acc-multi-agda}.  In \citep{alghunaim2024local}, LED achieves $\widetilde\cO(\frac{\kappa^2}{1-\sigma_2})$ communication complexity for the strongly convex setting with sufficient local steps. We improve this to $\widetilde\cO(\frac{\kappa}{1-\sigma_2})$ through the primal-dual interpretation of LED in Section~\ref{subsec:LED}. According to \citet{alghunaim2024local}, when the number of local steps is 1, LED reduces to Exact Diffusion \citep{yuan2018exact} and $\text{D}^2$ \citep{tang2018d}. 

\vspace{-2mm}
\paragraph{Comparison with Scaffnew/ProxSkip} When the momentum ($\beta$) is set to zero, Algorithm~\ref{alg:acc-multi-agda}
 closely resembles decentralized Scaffnew \citep{mishchenko2022proxskip}, with the primary difference that Scaffnew employs a random number of inner-loop iterations. We also detail a centralized version of Algorithm~\ref{alg:acc-multi-agda} in Algorithm~\ref{alg:cen-acc-multi-agda}, which is almost equivalent to centralized Scaffnew, as discussed in Section~\ref{sec:special_centralized}. Consequently, the control variate used in Scaffnew to mitigate client drift is interpreted as the dual variable in the Lagrangian formulation.

\begin{algorithm}[t]  
    \caption{Accelerated Gradient Ascent Multi-Stochastic gradient Descent (Acc-GA-MSGD)}
    \begin{algorithmic}[1] 
      \State \textbf{Input:} initial point $x_1^0, x_2^0, \dots, x_M^0$, and
     $\{\zeta_m^0\}_{m=1}^M$.
     \State \textbf{Initialize} $\widetilde \zeta^0_m = \zeta^0_m$ for $m=1, 2,\dots, M$
        \For{$t = 0,1,2,\dots, T$}
            \For{$k = 0,1,2,..., K-1$}
                 \State sample  $\{\xi^{k}_m\}_{m=1}^M$
                  \State  $x_m^{t, k+1} = x_m^{t, k} -\tau_1^k\left[\frac{1}{M}g_m(x_m^{t, k}; \xi_m^k)  + \widetilde\zeta_m^t  \right]$ for $m=1,2 ,\dots, M$
            \EndFor
            \State $x_m^{t+1, 0} = x_m^{t, K}$ for $m=1,2,\dots, M$ 
            \State $\zeta_m^{t+1} = \widetilde\zeta_m^t + \tau_2\left(x_m^{t+1, 0} - \sum_{j \in \cN_m}W_{m,j}x_j^{t+1,0} \right)$ for $m=1,2,\dots, M$
            \State $\widetilde\zeta_m^{t+1} = \zeta_m^{t+1} + \beta\left(\zeta_m^{t+1} - \zeta_m^{t} \right)$ for $m=1,2,\dots, M$
        \EndFor
        \State \textbf{Output:} $x_1^{T, K}, x_2^{T, K}, \dots, x_2^{T, K}$
    \end{algorithmic} 
\label{alg:acc-multi-agda}
\end{algorithm}

\subsection{Convergence Analysis for Decentralized Acc-GA-MSGD}

In contrast to our previous reformulation in the centralized setting, the coupling matrix between $X$ and $\lambda$ is no longer full-rank. However, we observe the following structure: $\lambda^t$ remains in $\spn(U) = \{U\hat \lambda: \hat \lambda\in \bR^{M\times n} \}$ based on its update rule (\ref{eq:lambda-update}), as long as the initial $\lambda^0 \in \text{Span}(U)$. Additionally, according to Lemma 3.1 in \citep{shi2015extra}, an optimal $\lambda^*$ exists in Span($U$). In the following, we present an intermediate step.

\begin{lemma}
\label{prop:duality-decentralized}
If a function $f: \bR^{m \times n} \rightarrow \bR$ is $\nu$-strongly convex and $\ell$-smooth, then $h(\lambda) = \min_x f(x) + \langle \lambda, A x\rangle$ is  $\sigma_{\min}^2/\ell$-strongly concave in the subspace $\spn(A)$, where  $\sigma_{\min}$ is the smallest nonzero singular values of $A$. This means that with $\lambda_1, \lambda_2 \in \spn(A)$, we have
\vspace{-2mm}
\begin{equation*}
     \langle \nabla h(\lambda_2) - \nabla h(\lambda_1),  \lambda_1-\lambda_2 \rangle \geq  \frac{\sigma_{\min}^2}{\ell}\|\lambda_1 - \lambda_2\|^2.
\end{equation*}
\end{lemma}
\vspace{-2mm}

A similar step is also described in \citep{scaman2017optimal}, but it assumes the existence of the Hessian of the Fenchel conjugate of $f$. This can be met when $f$ is twice differentiable. 
The lemma can be viewed as a generalization of Proposition~\ref{prop:duality}. In Lemma~\ref{lemma:dual-function-decentralized}, we show that the dual function $\Psi(\lambda) \triangleq \min_{X}\cL(X, \lambda)$ is $\frac{2M}{\mu}$-smooth and $\frac{M(1-\sigma_2)}{L}$-strongly concave in $\text{Span}(U)$, which helps to establish of the following linear convergence.

\begin{theorem}[Outer-loop Complexity]
\label{thm:outer-loop-decentral}
Under Assumption~\ref{assum:global}, \ref{assume:W} and \ref{assum:sc}, we choose inner loop iterations $K$ large enough such that $\bE \|X^{t, K} - X^*(\lambda^t)\|^2 \leq \delta$, where $X^*(\lambda^t) = \argmin_{X} \cL(X, \lambda^t)$ is the optimal primal variable corresponding to $\lambda^t$. If we initialize $\zeta^0 = \lambda^0 = 0$, the averaged output $\bar x^T = \frac{1}{M}\sum_{m=1}^M x_m^{T, K}$ satisfies
\begin{align*}
    & \bE F(\bar x^T) - \min_x F(x) \leq \exp \left(-\frac{T\sqrt{\mu(1-\sigma_2)}}{2 \sqrt{2L}}\right) \frac{ML^3\|X_0 - X^*\|^2}{\mu^2(1-\sigma_2)}   + \frac{17L^{\frac{7}{2}}\delta}{M\mu^{\frac{5}{2}} (1-\sigma_2)^\frac{5}{2}},
\end{align*}
where \(X^* = (x^* \dots  x^*)^\top \) with \(x^* \) being the optimal solution to the original problem (\ref{objective}).
\end{theorem}

\begin{lemma}[Inner-loop Complexity]
\label{thm:inner-loop-decentral}
Choosing $\delta = \frac{M\mu^{\frac{5}{2}} (1-\sigma_2)^\frac{5}{2}\epsilon}{34L^{\frac{7}{2}}}$ in Theorem~\ref{thm:outer-loop-decentral}, with proper stepsize $\{\tau_1^k\}_k$, the number of inner loop iterations  (where $\kappa = L/\mu$ is the condition number)
\begin{equation*}
     K = \cO\left(\kappa\log\left( \frac{\kappa L M\|X^0 - X^*\|^2}{(1-\sigma_2)\epsilon} + \kappa \right) + \frac{L^{\frac{7}{2}}\sigma^2}{\mu^{\frac{9}{2}}(1-\sigma_2)^{\frac{5}{2}}\epsilon} \right).
\end{equation*}
\end{lemma}

\begin{corollary} [Total Complexity]
\label{coro:total-complexity-decentral}
   Under Assumption~\ref{assum:global}, \ref{assume:W} and \ref{assum:sc}, to find an $\epsilon$-optimal solution $(x_1, x_2, \dots, x_M)$ such that $F(\bar x) - F^* \leq \epsilon$ and $\frac{1}{M}\sum_{m=1}^M \bE\|x_m - \bar x\|^2 \leq \cO(M\epsilon/L)$ with $\bar x\triangleq \frac{1}{M}\sum_{m=1}^M x_m$, the total number of communication rounds in Algorithm \ref{alg:acc-multi-agda} is 
   \begin{equation*}
       T = \cO \left( \sqrt{\frac{\kappa}{1-\sigma_2}} \log\left( \frac{\kappa LM\|X^0 - X^*\|^2}{\epsilon}\right)\right).
   \end{equation*}
   The total number of local gradients (samples) used by each client is 
   \begin{equation*}
       TK = \cO\left(\frac{\kappa^{\frac{3}{2}}}{\sqrt{1-\sigma_2}}\log^2\left( \frac{\kappa L M\|X^0 - X^*\|^2}{(1-\sigma_2)\epsilon} + \kappa \right) + \frac{\kappa^{4}\sigma^2}{\mu(1-\sigma_2)^{3}\epsilon}\log\left( \frac{\kappa LM\|X^0 - X^*\|^2}{\epsilon}\right) \right).
   \end{equation*}
\end{corollary}

\begin{remark}
\label{remark:lower-bound}
Acc-GA-MSGD achieves $\widetilde\cO\left(\sqrt{\frac{\kappa}{1-\sigma_2}}\right)$ communication complexity, which matches the lower bound for the deterministic setting ~\citep{scaman2017optimal, scaman2019optimal}.  Compared with other stochastic algorithms, our communication complexity has improved upon the $\widetilde\cO\left(\frac{\kappa^2}{1-\sigma_2}\right)$ complexity for LED \citep{alghunaim2024local}, and the $\widetilde\cO\left(\frac{\kappa}{(1-\sigma_2)c}\right)$ complexity for stochastic Gradient Tracking with large enough minibatch \citep{koloskova2021improved, pu2018distributed}, where $c > \Theta(1-\sigma_2)$ is another network-dependent constant.


\end{remark}

\subsection{Acceleration through Catalyst}

The Catalyst framework was adapted for distributed optimization in \citep{li2020revisiting}, detailed in Algorithm~\ref{alg:APPA-decentral}. In this adaptation, each subproblem involves all clients collaboratively solving a distributed problem, where the local function for each client is defined as \( F^s(x) = F_m(x) + L \left\|x - y_m^s\right\|^2 \). Further details are discussed in Section~\ref{subsec:cata-decentral}.

\begin{corollary}
\label{coro:catalyst-acc}
    If we solve the subproblem  in Catalyst with Algorithm~\ref{alg:acc-multi-agda}, the communication complexity and sample complexity are as follows:
    \begin{itemize}[leftmargin=*, itemsep=0pt]
\item \textbf{(SC) }  comm: $\widetilde\cO\left(\sqrt{\frac{\kappa}{1-\sigma_2}}\right)$; \ sample: $\widetilde\cO\left(\sqrt{\frac{\kappa}{1-\sigma_2}} + \frac{\sqrt \kappa\sigma^2}{\mu(1-\sigma_2)^3\epsilon}\right)$.
\item \textbf{(C) }  comm: $\widetilde\cO\left(\frac{1}{\sqrt{(1-\sigma_2)\epsilon}}\right)$; \ sample: $\widetilde\cO\left(\frac{1}{\sqrt{(1-\sigma_2)\epsilon}} + \frac{\sigma^2}{(1-\sigma_2)^3\epsilon^{2.5+\gamma/2}}\right)$ for $\gamma \in (0,1]$.
\item \textbf{(NC) }   comm: $\widetilde\cO\left(\frac{1}{\sqrt{1-\sigma_2}\epsilon^{2}}\right)$; \ sample: $\widetilde\cO\left(\frac{1}{\sqrt{1-\sigma_2}\epsilon^{2}} + \frac{\sigma^2}{(1-\sigma_2)^3\epsilon^{4}}\right)$.
\end{itemize}
\end{corollary}

\begin{remark}
    The communication complexities in strongly convex, convex, and nonconvex settings match with the lower bounds in \citep{scaman2017optimal, scaman2019optimal, lu2021optimal}, respectively, up to logarithmic terms. Although the sample complexities include the dependency in $1-\sigma_2$, it can be eliminated by Chebyshev acceleration, detailed in Section 4.2 of \citep{scaman2017optimal} and widely adopted in the literature~\citep{xu2020accelerated, kovalev2020optimal}. In the stochastic convex setting, our communication complexity improves over the  \(\cO((1-\sigma_2)^{-1}\epsilon^{-1})\) complexity in \citep{alghunaim2024local}. In the stochastic nonconvex setting, the existing near-optimal algorithm DeTAG~\citep{lu2021optimal} requires large minibatches.
\end{remark}



\section{Conclusion and Future Directions}
This work demonstrates that communication complexities comparable to those in deterministic settings can be achieved by stochastic local methods without using minibatches of large size. The connection between the primal-dual algorithms and local steps paves the way for applying extensive minimax optimization literature, including Gradient Descent Ascent (GDA)-type algorithms \citep{lin2020gradient, yang2020global, yang2022faster, boct2023alternating, zhang2022near, yan2020optimal}, to distributed optimization. Furthermore, it is intriguing to investigate the potential benefits of allowing each client to adopt adaptive optimization algorithms for distributed optimization within the primal-dual framework.

\section*{Acknowledgement}
This material is based upon work supported by Laboratory Directed Research and Development (LDRD) funding from Argonne National Laboratory, provided by the Director, Office of Science, of the U.S. Department of Energy under Contract No. DE-AC02-06CH11357. This work was also partially supported by the U.S. Department of Energy \textbf{Advanced Grid Modeling Program} under Grant DE-OE0000875. Murat Yildirim acknowledges support from the National Science Foundation under Grant No. 2104455.

\clearpage

\bibliography{ref}
\bibliographystyle{plainnat}


\clearpage
\appendix

\section{Proofs for Preliminaries}

\paragraph{Proof for Theorem \ref{thm:catalyst}}
\mbox{}

\begin{proof}
The convergence guarantees in the strongly convex and convex settings follow directly from Proposition 5 in \citep{lin2018catalyst}. We replace their deterministic stopping criterion with a stochastic equivalent with expectation, which ensures that the convergence guarantee also holds in expectation. The proof for this modification is almost the same, so we do not repeat it here. 

In the nonconvex setting, we present the proof here. Denote $$x_s^* = \argmin_x \left[ F^s(x) \triangleq F(x) + L\|x - x^s\|^2\right].$$ 
Because $F^s(x)$ is $L$-strongly convex, 
\begin{align} \nonumber
    \bE \|x^s - x_s^*\|^2 & \leq \frac{2}{L} \bE \left[F^s(x^s) - F^s(x_s^*)\right] \\ \nonumber
    &  \leq \frac{2}{L}\bE\left[F^s(x^s) - F^s(x^{s+1}) + \epsilon^s\right] \\ \nonumber
    & = \frac{2}{L}\bE\left[F(x^s) - F(x^{s+1}) - L\|x^{s+1} - x^s\|^2 + \epsilon^s\right] \\ \label{eq:nc-telescope}
    & \leq \frac{2}{L}\bE\left[F(x^s) - F(x^{s+1})  + \epsilon^s\right].
\end{align}
Since $\nabla F^s(x_s^*) = 0$, 
\begin{align*}
    \|\nabla F(x^{s+1})\| & = \|\nabla F(x^{s+1}) - \nabla F^s(x_s^*)\|  \\
    & = \|\nabla F(x^{s+1}) - \nabla F(x_s^*) + 2L(x_s^* - x^s) \|.
\end{align*}
Taking expectation, 
\begin{align*}
    \bE \|\nabla F(x^{s+1})\|^2 & = \bE  \|\nabla F(x^{s+1}) - \nabla F(x_s^*) + 2L(x_s^* - x^s) \|^2 \\
    & \leq 2\bE \|\nabla F(x^{s+1}) - \nabla F(x_s^*)\|^2 + 8L^2 \bE \|x_s^* - x^s\|^2 \\
    & \leq 2L^2 \bE\|x^{s+1} -x_s^*\|^2 + 8L^2 \bE \|x_s^* - x^s\|^2 \\
    & \leq 4L\epsilon^{s+1} + 8L^2 \bE \|x_s^* - x^s\|^2 .
\end{align*}
Combining with (\ref{eq:nc-telescope}) and telescoping,
\begin{align*}
    \sum_{s=1}^{S+1} \bE \|\nabla F(x^s)\|^2 \leq 4L\sum_{s=1}^{S+1} \epsilon^s  + 16L\Delta + 4L\sum_{s=0}^{S} \epsilon^s.
\end{align*}

\end{proof}

\begin{prop}
\label{prop:new-smooth-sc}
If each $F_m$ is $L$-smooth, i.e., $\|F_m(x) - F_m(\hat x)\| \leq L \|x - \hat x\|$ for all $x$ and $\hat x$, then \(H(X)= \frac{1}{M}\sum_{i=1}^M F_m\left(x_m\right)\) is $(L/M)$-smooth in $X$. If each  $F_m$ is $\mu$-strongly convex, then $H(X)$ is $(\mu/M)$-strongly convex  in $X$ . 
\end{prop}

\begin{proof}
When each $F_m$ is $L$-smooth, 
\begin{align*}
\| \nabla H(X) - \nabla H(\hat X)\|^2 & = 
\left\| \frac{1}{M}\begin{pmatrix}
\nabla F_1(x_1)^\top\\
\nabla F_2(x_2)^\top\\
\vdots\\
\nabla F_M(x_M)^\top
\end{pmatrix} -
\frac{1}{M} \begin{pmatrix}
\nabla F_1(\hat x_1)^\top\\
\nabla F_2(\hat x_2)^\top\\
\vdots\\
\nabla F_M(\hat x_M)^\top
\end{pmatrix} \right\|^2\\
& = \frac{1}{M^2}\sum_m\left\| \nabla F_m(x_m) - \nabla F_m(\hat x_m)\right\|^2  \\
& \leq \frac{1}{M^2}\sum_m L^2\|x_m - \hat x_m\|^2 = \frac{L^2}{M^2} \|X -\hat X\|^2.
\end{align*}
Therefore, $H$ is $L/M$-smooth.

When each $F_m$ is $\mu$-strongly convex, 
\begin{align*}
\langle\nabla H(X) - \nabla H(\hat X), X - \hat X\rangle & = 
\Biggl<\frac{1}{M}\begin{pmatrix}
\nabla F_1(x_1)^\top - F_1(\hat x_1)^\top\\
\vdots\\
\nabla F_M(x_M)^\top - F_M(\hat x_M)^\top
\end{pmatrix}, 
\begin{pmatrix}
x_1^\top - \hat x_1^\top \\
\vdots\\
x_M^\top - \hat x_M^\top
\end{pmatrix} \Biggr> \\
& = \frac{1}{M}\sum_m\left[\nabla F_m(x_m) - F_m(\hat x_m) \right]^{\top } (x_m - \hat x_m) \\
& \geq \frac{1}{M}\sum_m \mu \|x_m - \hat x_m\|^2   = \frac{\mu}{M}  \|X -\hat X\|^2 .
\end{align*}
Therefore, $H$ is $\mu/M$-strongly convex. 

\end{proof}
\section{Proofs for the Centralized Setting}

\begin{lemma}
\label{lemma:dual-function}
Under Assumption ~\ref{assum:global} and \ref{assum:sc}, the dual function $\widetilde \Psi(\lambda) \triangleq \min_{x_1, X} \widetilde \cL(x_1, X, \lambda)$ is $\frac{4M}{\mu}$-smooth and $\frac{2M}{3L}$-strongly concave.
\end{lemma}

\begin{proof}
    We first note that the dual function is separable in $X$ and $x_1$:
\begin{align*} 
   \widetilde \Psi(\lambda)  & =  \min_{x_1, X} H(x_1, X) - \frac{\mu}{4M}\|X\|^2 + \frac{\mu(M-1)}{4M}\|x_1\|^2 + 
     \langle
     \lambda, X - \one x_1^\top \rangle\\
   & = \min_X \left[ \frac{1}{M} \sum_{m=2}^M F_m(x_m) - \frac{\mu}{4M}\|X\|^2 +
    \langle \lambda,  X \rangle \right] + \\
   & \fakeeq \min_{x_1}\left[ \frac{1}{M}F_1(x_1) + \frac{\mu(M-1)}{4M}\|x_1\|^2  - \langle \lambda, \one x_1\rangle\right]
  \end{align*}    
Define 
$$\Psi_X(\lambda) \triangleq \min_X \left[ \frac{1}{M} \sum_{m=2}^M F_m(x_m) - \frac{\mu}{4M}\|X\|^2 +
    \langle \lambda,  X \rangle \right].$$ 
Note that the function inside the bracket is $\mu/(2M)$-strongly convex and $(\mu+2L)/(2M)$-smooth in $X$ by Proposition~\ref{prop:new-smooth-sc}, and the coupling matrix between $\lambda$ and $X$ is the identity matrix. According to Proposition~\ref{prop:duality} implies $\Psi_X(\lambda)$ is $2M/\mu$-smooth and $2M/(\mu+2L)$-strongly concave. Define 
$$\Psi_{x_1}(\lambda) \triangleq \min_{x_1}\left[ \frac{1}{M}F_1(x_1) + \frac{\mu(M-1)}{4M}\|x_1\|^2  - \langle \lambda, \one x_1\rangle\right].$$ 
Note that the function inside the bracket is $(M+1)\mu/(2M)$-strongly convex and $((M-1)\mu + 2L)/(2M)$-smooth in $x_1$. The largest singular value of $\one$ is $\sqrt{M-1}$. According to Proposition~\ref{prop:duality} implies $\Psi_X(\lambda)$ is $2(M-1)M/((M+1)\mu)$-smooth and concave. Since $\widetilde\Psi(\lambda) = \Psi_X(\lambda) + \Psi_{x_1}(\lambda)$, we conclude by adding smoothness and strong concavity constant for two components together. 
\end{proof}

\begin{theorem}
\label{Thm:outer-loop}
Under Assumption~\ref{assum:global} and \ref{assum:sc}, we choose inner loop iterations $K$ large enough such that $\bE \sum_{m=2}^M \|x_m^{t, K} - x_m^*(\lambda^t)\|^2 \leq \delta_1$ and $\bE \|x_1^{t, K} - x_1^*(\lambda^t)\|^2 \leq \delta_2$, where $(x_1^*, x_2^*, \dots, x_M^*) = \argmin_{x_1, X} \widetilde\cL(x_1, X, \lambda^t)$ is the optimal primal variable corresponding to $\lambda^t$. With stepsize  $\tau_3 = \frac{\mu}{4M}$, the last iterate of Algorithm~\ref{alg:multi-agda}, $\lambda^T$, satisfies:
\begin{equation*}
    \bE \|\lambda^T - \lambda^*\|^2 \leq \frac{6L}{\mu}  \left(1 - \frac{\mu}{6L} \right)^T  \|\lambda^0 - \lambda^*\|^2 + \frac{9L^2\delta_1}{2M^2} + \frac{9L^2(M-1)\delta_2}{2M^2}.
\end{equation*}
where $\lambda^*$ is the optimal dual solution to $\widetilde\cL(x_1, X, \lambda)$.

\end{theorem}

\begin{proof}
Because $\widetilde \cL(X, x_1, \lambda)$ is strongly convex in $x_1$ and $X$, by Danskin's Theorem, 
$$
\nabla \widetilde\Psi(\lambda) = X^*(\lambda) - \one (x_1^*(\lambda))^\top  \quad \text{ where } (x_1^*(\lambda), X^*(\lambda)) \triangleq \argmin_{x_1, X} \widetilde \cL(x_1, X, \lambda).
$$
Since $\widetilde \Psi$ is $\frac{4M}{\mu}$-smooth by Lemma~\ref{lemma:dual-function}, 
\begin{align*}
    -\widetilde\Psi(\lambda^{t+1}) & \leq -\widetilde\Psi(\lambda^{t}) - \langle \nabla \widetilde \Psi(\lambda^t), \lambda^{t+1} - \lambda^t\rangle + \frac{2M}{\mu}\|\lambda^{t+1} - \lambda^t\|^2 \\
    & = -\widetilde\Psi(\lambda^{t}) - \tau_3\langle \nabla \widetilde \Psi(\lambda^t), X^{t, K} - \one (x_1^{t,K})^\top \rangle + \frac{2M\tau_3^2}{\mu}\|X^{t, K} - \one (x_1^{t,K})^\top\|^2 \\
    & = -\widetilde\Psi(\lambda^{t}) - \tau_3\langle \nabla \widetilde \Psi(\lambda^t), X^{t, K} - \one (x_1^{t,K})^\top \rangle + \frac{\tau_3}{2}\|X^{t, K} - \one (x_1^{t,K})^\top\|^2 \\
    & = -\widetilde\Psi(\lambda^{t}) - \frac{\tau_3}{2}\| \nabla \widetilde \Psi(\lambda^t)\|^2 + \frac{\tau_3}{2}\|X^{t, K} - \one (x_1^{t,K})^\top - \nabla \widetilde \Psi(\lambda^t)\|^2,
\end{align*}
where in the second equality, we use $\tau_3 = \frac{\mu}{4M}$. Defining $\Psi^* = \max_{\lambda} \widetilde \Psi(\lambda)$, 
\begin{align} \nonumber
    \Psi^* -\widetilde\Psi(\lambda^{t+1}) & \leq \Psi^* -\widetilde\Psi(\lambda^{t}) - \frac{\tau_3}{2}\| \nabla \widetilde \Psi(\lambda^t)\|^2 + \frac{\tau_3}{2}\|X^{t, K} - \one (x_1^{t,K})^\top - \nabla \widetilde \Psi(\lambda^t)\|^2 \\ \nonumber
    & \leq \Psi^* -\widetilde\Psi(\lambda^{t}) - \frac{\mu}{6L} [\Psi^* -\widetilde\Psi(\lambda^{t}) ] + \frac{\tau_3}{2}\|X^{t, K} - \one (x_1^{t,K})^\top - \nabla \widetilde \Psi(\lambda^t)\|^2 \\ \label{eq:ga-psi}
    & = \left(1 - \frac{\mu}{6L} \right) [\Psi^* -\widetilde\Psi(\lambda^{t})] + \frac{\mu}{8M}\|X^{t, K} - \one (x_1^{t,K})^\top - \nabla \widetilde \Psi(\lambda^t)\|^2,
\end{align}
where in the second inequality, we use the strong concavity of $\widetilde \Psi$. Note that 
\begin{align*}
    \bE \|X^{t, K} - \one (x_1^{t,K})^\top - \nabla \widetilde \Psi(\lambda^t)\|^2 & = \bE \|X^{t, K} - \one (x_1^{t,K})^\top - X^*(\lambda) + \one (x_1^*(\lambda))^\top \|^2 \\
    & \leq 2 \bE \|X^{t, K}  - X^*(\lambda) \|^2 + 2 \bE \|\one (x_1^{t,K})^\top - \one (x_1^*(\lambda))^\top \|^2 \\
    &  \leq 2 \bE \|X^{t, K}  - X^*(\lambda) \|^2 + 2(M-1) \bE \| x_1^{t,K} -  x_1^*(\lambda)\|^2 \\
    & \leq 2\delta_1^2 + 2(M-1)\delta_2^2.
\end{align*}
Combining with (\ref{eq:ga-psi}), 
\begin{equation*}
    \bE \left[\Psi^* -\widetilde\Psi(\lambda^{t+1}) \right] \leq \left(1 - \frac{\mu}{6L} \right)^t \bE \left[\Psi^* -\widetilde\Psi(\lambda^{t})\right] + \frac{\mu \delta_1}{4M} + \frac{\mu(M-1)\delta_2}{4M}.
\end{equation*}
Recursing it, 
\begin{equation*}
    \bE \left[\Psi^* -\widetilde\Psi(\lambda^{t}) \right] \leq \left(1 - \frac{\mu}{6L} \right)^t \left[\Psi^* -\widetilde\Psi(\lambda^{0})\right] + \frac{3L\delta_1}{2M} + \frac{3L(M-1)\delta_2}{2M}.
\end{equation*}
Further with $\frac{2M}{3L}$-strong concavity  and $\frac{4M}{\mu}$-smoothness of $\widetilde \Psi$, 
\begin{align*}
    \bE \|\lambda^t - \lambda^*\|^2 \leq \frac{3L}{M}  \left(1 - \frac{\mu}{6L} \right)^t \ \left[\Psi^* -\widetilde\Psi(\lambda^{0})\right] + \frac{9L^2\delta_1}{2M^2} + \frac{9L^2(M-1)\delta_2}{2M^2} \\
    \leq \frac{6L}{\mu}  \left(1 - \frac{\mu}{6L} \right)^t  \|\lambda^0 - \lambda^*\|^2 + \frac{9L^2\delta_1}{2M^2} + \frac{9L^2(M-1)\delta_2}{2M^2}.
\end{align*}

\end{proof}

\paragraph{Proof for Theorem \ref{thm:outer-loop-central}}
\mbox{}

\begin{proof}
Note that 
\begin{align*}
    & \fakeeq \bE \|X^{T, K} - X^*\|^2 + \bE \|x_1^{T, K} - x_1^*\|^2 \\ 
    & \leq 2\bE \left[\|X^{T, K}-X^*(\lambda^T)\|^2 + \|X^*(\lambda^T) - X^*\|^2\right] +  2\bE\left [\|x_1^{T, K}-x_1^*(\lambda^T)\|^2 + \|x_1^*(\lambda^T) - x_1^*\|^2\right]\\
    & \leq 2\delta_1 + 2\bE\|X^*(\lambda^T) - X^*(\lambda^*)\|^2 + 2\delta_2 + 2\bE \|x_1^*(\lambda^T) - x_1^*(\lambda^*)\|^2.
\end{align*}
According to Lemma \ref{lemma:dual-function}, since $\widetilde \cL(x_1, X, \lambda)$ is $\frac{\mu}{2M}$-strongly convex in $X$ and the coupling matrix between $X$ and $\lambda$ has the largest singular value 1, we know $X^*(\lambda)$ is $\frac{2M}{\mu}$-Lipschitz. Since    $\widetilde \cL(x_1, X, \lambda)$ is  $\frac{(M+1)\mu}{2M}$-strongly convex  in $x_1$ and the coupling matrix $\one$ has the largest singular value $\sqrt{M-1}$, we know $x_1^*(\lambda)$ is $\frac{2M}{\mu\sqrt{M+1}}$-Lipschitz. Therefore
\begin{equation*}
    \bE \|X^{T, K} - X^*\|^2 + \bE \|x_1^{T, K} - x_1^*\|^2 \leq 2(\delta_1 +\delta_2) + \frac{8M^2}{\mu^2}\bE \|\lambda^T - \lambda^*\|^2  + \frac{8M^2}{\mu^2(M+1)}\bE \|\lambda^T - \lambda^*\|^2.
\end{equation*}
Note that the optimal solution subproblem $(x_1^*, X^*)$ should satisfy that all clients have the same variable, i.e., $ (x_1^*, X^*) = (x^*, x^*,..., x^*)$ for some $x^*$. This is because the solution to $\min_{x_1, X}\max_{\lambda} \widetilde L(x_1, X, \lambda)$ must be unique due to $\widetilde \Psi$ being strongly concave and $\widetilde L(x_1, X, \lambda)$ being strongly convex in $x_1$ and $X$. If $(x_1^*, X^*)$ does not have the same copy for all clients, then $\lambda^*$ does not maximize $\widetilde L(x_1^*, X^*, \lambda)$, which leads to a contradiction. With the definition of $
\bar x^T = \frac{1}{M}\sum_{m=1}^M x_m^{T, K}$,
by Jensen's inequality,
$$
\bE \|\bar x^T -x^*_s\|^2 \leq \frac{1}{M}\sum_{m=1}^M\bE\|x_m^{T, K} - x^*\|^2 \leq \frac{2(\delta_1 + \delta_2)}{M} + \frac{8(M+1)}{\mu^2}\bE \|\lambda^T - \lambda^*\|^2.
$$
This $x^*$ should also be the unique solution to $\min_{x}F(x)$ because the primal function $\max_{\lambda} \widetilde L(x_1, X, \lambda) = F(x)$. Since $F(x)$ is $L$-smooth, 
\begin{align} \nonumber
    \bE F(\bar x^T) - \min_x F(x) & \leq \frac{L}{2}\bE \|\bar x^T - x^*\|^2 \\ \nonumber
    & \leq \frac{L(\delta_1 + \delta_2)}{M} + \frac{4L(M+1)}{\mu^2}\bE \|\lambda^T - \lambda^*\|^2   \\ \nonumber
    &\leq \frac{L(\delta_1 + \delta_2)}{M} + \frac{4L(M+1)}{\mu^2} \left[\frac{6L}{\mu}  \left(1 - \frac{\mu}{6L} \right)^t  \|\lambda^0 - \lambda^*\|^2 + \frac{9L^2\delta_1}{2M^2} + \frac{9L^2(M-1)\delta_2}{2M^2}\right] \\ \label{eq:xbar-bound}
    & \leq \frac{24L^2(M+1)}{\mu^3}  \left(1 - \frac{\mu}{6L} \right)^t  \|\lambda^0 - \lambda^*\|^2   + \frac{40L^3}{\mu^2 M} \delta_1 + \frac{40L^3}{\mu^2 } \delta_2,
\end{align}
where in the third inequality we use Theorem~\ref{Thm:outer-loop}. By the optimality of $x_m^*$ and $\lambda^*$, i.e., $\nabla_{x_m}\widetilde L(x_1, X, \lambda)\vert_{x_1 = x_1^*, X = X^*, \lambda = \lambda^*} = 0$, we have
\begin{equation*}
    \lambda^*_m = -\frac{1}{M}\nabla F_m(x^*)  + \frac{L}{2M}x^*
\end{equation*}
With our initialization of $\lambda_m^0$, 
\begin{align*}
    \bE\left\|\lambda_m^0 - \left(- \frac{1}{M}\nabla F_m(x^0)  + \frac{L}{2M}x^0 \right)\right\|^2 \leq \frac{\sigma^2}{M}.
\end{align*}
Therefore, we can bound
\begin{align} \nonumber
    \bE\|\lambda_m^0 - \lambda_m^*\|^2 & \leq  2\bE\left\|\lambda_m^0 - \left(- \frac{1}{M}\nabla F_m(x^0)  + \frac{L}{2M}x^0 \right)\right\|^2 +   2\bE\left\|\lambda_m^* - \left(- \frac{1}{M}\nabla F_m(x^0)  + \frac{L}{2M}x^0 \right)\right\|^2 \\ \nonumber
    & = \frac{2\sigma^2}{M} + 2\left\|-\frac{1}{M}\left[\nabla F_m(x^0) - \nabla F_m(x^*)\right] - \frac{L}{2M}\left(x^0 - x^* \right) \right\|^2 \\ \nonumber
    & \leq \frac{2\sigma^2}{M} + \frac{5L^2}{M^2}\|x^0 - x^*\|^2,
\end{align}
where in the second inequality we use the smoothness of $F_m$. Therefore, 
\begin{equation} \label{eq:initial-lambda-bd}
    \bE\|\lambda^0 - \lambda^*\|^2 = \sum_{m=1}^M \bE\|\lambda_m^0 - \lambda_m^*\|^2 \leq 2\sigma^2 + \frac{5L^2}{M}\|x^0 - x^*\|^2.
\end{equation}
Plugging back into (\ref{eq:xbar-bound}),
\begin{equation*}
     \bE F(\bar x^T) - \min_x F(x) \leq  \frac{48L^4}{\mu^3}  \left(1 - \frac{\mu}{6L} \right)^t  \left[\|x^0-x^*\|^2 + \frac{2\sigma^2M}{L^2} \right]   + \frac{40L^3}{\mu^2 M} \delta_1 + \frac{40L^3}{\mu^2 } \delta_2.
\end{equation*}
\end{proof}

\paragraph{Proof for Lemma \ref{thm:inner-loop-central}}
\mbox{}

\begin{proof}
    $\widetilde \cL(x_1, X, \lambda)$ is $\mu/(2M)$-strongly convex and $(\mu+2L)/(2M)$-smooth in $X$. Also, we have access to a stochastic gradient estimate for $\nabla_X \widetilde \cL(x_1, X, \lambda)$ with variance $\frac{\sigma^2}{M}$ because
$$ \mathbb{E} \left\| \frac{1}{M}\begin{pmatrix}
\nabla F_2(x_2)\\
\nabla F_3(x_3)\\
\vdots\\
\nabla F_M(x_M)
\end{pmatrix} -
\frac{1}{M} \begin{pmatrix}
g_2(x_2;\xi_2)\\
g_3(x_3; \xi_3)\\
\vdots\\
g_M(x_M;\xi_M)
\end{pmatrix} \right\|^2 \leq \frac{\sigma^2}{M}.$$
By the convergence rate of SGD on strongly convex and smooth functions (e.g., Theorem 5 of \citep{stich2019unified}), to achieve $\bE\|X^{t, K} - X^*(\lambda^t)\|^2 \leq \delta_1 = \frac{\mu^2 M \epsilon}{120L^3}$,
\begin{equation} \label{eq:inner-loop-bd}
    K = \cO\left(\kappa\log\left(\frac{L^4\bE\|X^{t,0} -  X^*(\lambda^t)\|^2}{\mu^3M\epsilon} \right) + \frac{L^3\sigma^2}{\mu^4\epsilon} \right).
\end{equation}
Now we bound $\bE\|X^{t,0} -  X^*(\lambda^t)\|^2$. When $t \geq 1$, 
\begin{align} \nonumber
    \bE\|X^{t,0} -  X^*(\lambda^t)\|^2 & \leq 2\bE\|X^{t,0} -  X^*(\lambda^{t-1})\|^2 + 2\bE\|X^*(\lambda^{t-1}) -  X^*(\lambda^t)\|^2 \\ \nonumber
    & \leq 2\bE\|X^{t,0} -  X^*(\lambda^{t-1})\|^2 + \frac{8M^2}{\mu^2} \bE\|\lambda^{t-1}- \lambda^t\|^2 \\  \label{eq:initial-x0}
    & \leq \frac{\mu^2 M \epsilon}{60L^3} + \frac{8M^2}{\mu^2} \left[ \bE\|\lambda^{t-1}- \lambda^*\|^2 + \bE\|\lambda^{t}- \lambda^*\|^2  \right],
\end{align}
where we use Proposition~\ref{prop:duality} in the second inequality.  According to Theorem \ref{Thm:outer-loop}, for all $t \geq 1$,
\begin{align*}
    \bE \|\lambda^t - \lambda^*\|^2 & \leq \frac{6L}{\mu}  \left(1 - \frac{\mu}{6L} \right)^t  \bE\|\lambda^0 - \lambda^*\|^2 + \frac{9L^2\delta_1}{2M^2} + \frac{9L^2(M-1)\delta_2}{2M^2} \\
    & \leq \frac{6L}{\mu} \bE \|\lambda^0 - \lambda^*\|^2 + \cO\left( \frac{\mu^2\epsilon}{LM}\right) \\
    & \leq \frac{12L\sigma^2}{\mu } + \frac{30L^3}{\mu M}\|x^0 - x^*\|^2   + \cO\left( \frac{\mu^2\epsilon}{LM}\right).
\end{align*}
where in the last inequality we use (\ref{eq:initial-lambda-bd}). Therefore, 
\begin{equation*}
     \bE\|X^{t,0} -  X^*(\lambda^t)\|^2 \leq \cO\left(\frac{\mu^2M\epsilon}{L^3} + \frac{ML^3}{\mu^3}\|x^0 - x^*\|^2 + \frac{M^2L\sigma^2}{\mu^3} \right).
\end{equation*}
Combining with (\ref{eq:inner-loop-bd}), 
\begin{equation*}
     K = \cO\left(\kappa\log\left(\frac{ML^5\sigma^2}{\mu^6 \epsilon} + \frac{L^7\|x^0 - x^*\|^2}{\mu^6\epsilon} + \frac{L}{\mu} \right) + \frac{L^3\sigma^2}{\mu^4\epsilon} \right).
\end{equation*}
When $t = 0$, 
\begin{align*}
    \|X^{0,0} -  X^*(\lambda^0)\|^2 & \leq 2 \|X^{0,0} -  X^*\|^2 + 2\|X^* -  X^*(\lambda^0)\|^2 \\
    & \leq 2 (M-1)\|x^0 - x^*\|^2 + \frac{8M^2}{\mu^2}\|\lambda^0 - \lambda^*\|^2 \\
    & \leq 2 (M-1)\|x^0 - x^*\|^2 + \frac{16M^2\sigma^2}{\mu^2} + \frac{40L^2M\|x^0 - x^*\|^2}{\mu^2}.
\end{align*}
Combining with (\ref{eq:inner-loop-bd}), 
\begin{equation*}
     K = \cO\left(\kappa\log\left(\frac{ML^4\sigma^2}{\mu^5 \epsilon} + \frac{L^6\|x^0 - x^*\|^2}{\mu^5\epsilon} \right) + \frac{L^3\sigma^2}{\mu^4\epsilon} \right).
\end{equation*}
Now we finish the proof for the required number of inner-loop iterations to achieve $\bE \sum_{m=2}^M \|x_m^{t, K} - x_m^*(\lambda^t)\|^2 \leq \delta_1$. Similar proof can be used to show the inner-loop complexity for achieving  $\bE \|x_1^{t, K} - x_1^*(\lambda^t)\|^2 \leq \delta_2$.

\end{proof}

\subsection{Catalyst Acceleration}

\paragraph{Proof for Corollary \ref{coro:catalyst-GA-MSGD}}
\mbox{}

\begin{proof}
After adding the regularization, the subproblem (\ref{cata:subproblem}) in Catalyst is $(\mu+2L)$-strongly convex and $3L$-smooth. Therefore, the condition number of the subproblems is $\cO(1)$. According to Corollary~\ref{coro:total-complexity}, the communication complexity for solving subproblem at iteration $s$ is 
   \begin{equation}
   \label{eq:Ts}
       T_s = \widetilde\cO \left(\log\left( \frac{L\|x^0_s - x_s^*\|^2}{\epsilon^s} + \frac{M\sigma^2}{L\epsilon^s} \right)\right),
   \end{equation}
   and the total number of local gradients (samples) used by each client is 
   \begin{equation}
   \label{eq:TsKs}
       T_sK_s = \widetilde\cO \left(\log^2\left(\frac{M \sigma^2}{\mu \epsilon^s} + \frac{ L\|x_s^0 - x_s^*\|^2}{\epsilon^s} + \frac{1}{\epsilon^s}  \right) + \frac{\sigma^2}{L\epsilon^s}  \log\left( \frac{ L\|x_s^0 - x_s^*\|^2}{\epsilon^s} + \frac{M\sigma^2}{L\epsilon^s} \right)\right),
   \end{equation}
where $x_s^0$ denotes the initial point of GA-MSGD in solving $s$-th subproblem, $x_s^*$ denotes the optimal solution for $s$-th subproblem, and $\epsilon^s$ is the target accuracy.

\paragraph{Strongly Convex.} By Theorem~\ref{thm:catalyst}, the number of Catalyst iterations to achieve $\bE F(x^S) - F^* \leq \epsilon$ is 
$$
S = \widetilde\cO(\sqrt{\kappa}),
$$
here $\kappa$ is the condition number of the original objective. Now we will bound the subproblem target accuracy. From Theorem~\ref{thm:catalyst},  
\begin{equation*}
    \bE F\left(x^S\right)-F^* \leq \frac{800(1-0.9\sqrt{q})^{S+1}\Delta}{q}  = \epsilon 
      \ \Longleftrightarrow \  (1-0.9\sqrt{q})^{S+1}\Delta= \frac{q\epsilon}{800}.
\end{equation*}
Therefore, for $s \leq S$,
\begin{equation*}
    \epsilon^s = \frac{2(1-0.9\sqrt{q})^s\Delta}{9} \geq \frac{2(1-0.9\sqrt{q})^{S+1}\Delta}{9} = \frac{q\epsilon}{3600} = \Theta\left( \frac{\epsilon}{\kappa}\right).
\end{equation*}
By Proposition 12 in \citep{lin2018catalyst}, if we pick $x_s^0 = x^{s-1} + \frac{2L}{2L + \mu}(y^s - y^{s-1})$, then the distance $\bE\|x_s^0 - x_s^*\|^2$ is bounded. Plugging back into (\ref{eq:Ts}) and (\ref{eq:TsKs}), we have $T_s = \widetilde \cO\left(1\right)$ and $T_sK_s = \widetilde \cO\left(1 + \frac{\kappa\sigma^2}{L\epsilon} \right)$. The total communication rounds is $\sum_{s}T_s = \widetilde \cO\left(\sqrt \kappa\right)$, and total sample complexity for each client is $\sum_{s}T_sK_s = \widetilde \cO\left(\sqrt \kappa + \frac{\kappa^{1.5}\sigma^2}{L\epsilon}\right) =  \widetilde \cO\left(\sqrt \kappa + \frac{\kappa^{1/2}\sigma^2}{\mu\epsilon}\right)$.

\paragraph{Convex. }  By Theorem~\ref{thm:catalyst}, the number of Catalyst iterations to achieve $\bE F(x^S) - F^* \leq \epsilon$ is 
$$
S = \widetilde\cO\left(\sqrt{\frac{L\|x_0 - x^*\|^2}{\gamma^2\epsilon}} \right).
$$
Now we will bound the subproblem target accuracy. From Theorem~\ref{thm:catalyst}, we can pick $S$ such that 
\begin{equation*}
    \bE F\left(x^S\right)-F^* \leq \frac{8}{(S+1)^2} \cdot \frac{4\Delta}{\gamma^2} = \frac{\epsilon}{2} 
      \ \Longleftrightarrow \  (S+1)^2 = \frac{64\Delta}{\gamma^2\epsilon}.
\end{equation*}
Therefore, for $s \leq S$,
\begin{equation*}
    \epsilon^s = 
\frac{2 \Delta}{9(s+1)^{4+\gamma}}
 \geq \frac{2 \Delta}{9(s+1)^{4+\gamma}} = \Theta\left( \frac{\gamma^{4+\gamma}\epsilon^{2+\frac{\gamma}{2}}}{\Delta^{1+\frac{\gamma}{2}}}\right).
\end{equation*}
By Proposition 12 in \citep{lin2018catalyst}, if we pick $x_s^0 = x^{s-1} + \frac{2L}{2L + \mu}(y^s - y^{s-1})$, then the distance $\bE\|x_s^0 - x_s^*\|^2$ is bounded. Plugging back into (\ref{eq:Ts}) and (\ref{eq:TsKs}), we have $T_s = \widetilde \cO\left(1\right)$ and $T_sK_s = \widetilde \cO\left(1 + \frac{\Delta^{1+\frac{\gamma}{2}}\sigma^2}{L\gamma^{4+\gamma}\epsilon^{2+\frac{\gamma}{2}}} \right)$. The total communication rounds is $\sum_{s}T_s = \widetilde \cO\left(\sqrt{\frac{L\|x_0 - x^*\|^2}{\epsilon}}\right)$, and total sample complexity for each client is $\sum_{s}T_sK_s = \widetilde \cO\left(\sqrt{\frac{L\|x_0 - x^*\|^2}{\epsilon}} + \frac{\Delta^{1+\frac{\gamma}{2}}\sigma^2\|x_0 - x^*\|^2}{\sqrt{L}\gamma^{4+\gamma}\epsilon^{2.5+\frac{\gamma}{2}}}\right)$.

\paragraph{Nonconvex. }  By Theorem~\ref{thm:catalyst}, the number of Catalyst iterations to achieve $\bE\|\nabla F(x)\|^2 \leq \epsilon^2$ is 
$$
S = \widetilde\cO\left(\frac{L\Delta}{\epsilon^2}\right).
$$
Now we will bound the subproblem target accuracy. From Theorem~\ref{thm:catalyst}, we pick $S$ such that 
\begin{equation*}
\frac{1}{S} \sum_{s=1}^{S} \mathbb{E}\left\|\nabla F\left(x^s\right)\right\|^2 \leq \frac{32 L \Delta}{S} = \epsilon^2
      \ \Longleftrightarrow \  S = \frac{32 L \Delta}{\epsilon^{2}}.
\end{equation*}
Therefore, 
\begin{equation*}
    \epsilon^s = 
\frac{\Delta}{S} = \frac{\epsilon^2}{32L}.
\end{equation*}
The distance $\bE\|x_s^0 - x_x^*\|^2$ is bounded, because by (\ref{eq:nc-telescope}), if we pick $x_s^0$ to be last iterate in Catalyst $x^s$
\begin{align} \nonumber
    \bE\|x_s^0 - x_x^*\|^2 =  \bE \|x^s - x_s^*\|^2 & \leq \frac{2}{L}\bE\left[F(x^s) - F(x^{s+1})  + \epsilon^s\right] \\ \nonumber
    & \leq \frac{2}{L}\sum_{s =0}^S\bE\left[F(x^s) - F(x^{s+1})  + \epsilon^s\right] \leq \frac{4\Delta}{L}.
\end{align}
Plugging back into (\ref{eq:Ts}) and (\ref{eq:TsKs}), we have $T_s = \widetilde \cO\left(1\right)$ and $T_sK_s = \widetilde \cO\left(1 + \frac{\sigma^2}{\epsilon^2}\right)$. The total communication rounds is $\sum_{s}T_s = \widetilde \cO\left(\frac{L\Delta}{\epsilon^2}\right)$, and total sample complexity for each client is $\sum_{s}T_sK_s = \widetilde \cO\left(\frac{L\Delta}{\epsilon^2} +  \frac{L\Delta\sigma^2}{\epsilon^4}\right)$.

\end{proof}

\paragraph{Proof for Corollary \ref{coro:catalyst-scaffold}}
\mbox{}

\begin{proof}
After adding the regularization, the subproblem (\ref{cata:subproblem}) in Catalyst is $(\mu+2L)$-strongly convex and $3L$-smooth. Therefore, the condition number of the subproblems is $\cO(1)$. According to Theorem~\ref{thm:scaffold}, the communication complexity for solving subproblem at iteration $s$ is (we hide the dependency on parameters other than $\epsilon$ and 
$\kappa$ in $\widetilde\cO$ below)
   \begin{equation*}
       T_s = \widetilde\cO \left(1 \right),
   \end{equation*}
   and the total number of local gradients (samples) used by each client is 
   \begin{equation*}
       T_sK_s = \widetilde\cO \left(1 + \frac{\sigma^2}{LM\epsilon^s}\right).
   \end{equation*}
The rest of the proof follows the same reasoning as the proof for Corollary~\ref{coro:catalyst-GA-MSGD}.

\end{proof}
\section{Proofs for the Decentralized Setting}

\subsection{Useful Lemmas}

\bigskip

\paragraph{Proof for Proposition \ref{prop:duality} and Lemma \ref{prop:duality-decentralized}}
\mbox{}

\begin{proof} Here we use $\sigma_{\min}$ to denote smallest nonzero singular values of $A^\top$. We use Frobenius norm and the inner product $\langle A, B\rangle = Tr(A^\top B)$ by default. \\
\textbf{Smoothness:} Denote $x^*(\lambda) = \min_x f(x) + \langle \lambda, Ax\rangle$. By the optimality condition, for any $\lambda_1$ and $\lambda_2$, we have 
\begin{align*}
    & \langle x-x^*(\lambda_1), \nabla f(x^*(\lambda_1)) + A^\top \lambda_1\rangle \geq 0, \quad \forall x;\\
    & \langle x-x^*(\lambda_2), \nabla f(x^*(\lambda_2)) + A^\top \lambda_2 \rangle \geq 0, \quad \forall x.
\end{align*}
Plugging in $x= x^*(\lambda_2)$ in the first inequality and $x= x^*(\lambda_1)$ in the second inequality, and adding them,
\begin{equation}
\label{eq:prop_optimality}
    \langle x^*(\lambda_2)-x^*(\lambda_1), \nabla f(x^*(\lambda_1)) - \nabla f(x^*(\lambda_2)) + A^\top \lambda_1 -A^\top \lambda_2 \rangle \geq 0.
\end{equation}
Because $f$ is $\nu$-strongly convex,
\begin{equation*}
    \langle x^*(\lambda_2)-x^*(\lambda_1), \nabla f(x^*(\lambda_2)) - \nabla f(x^*(\lambda_1))  \rangle \geq \nu\|x^*(\lambda_2) - x^*(\lambda_1)\|^2.
\end{equation*}
Combining the previous two inequalities,
\begin{equation*}
     \langle x^*(\lambda_2)-x^*(\lambda_1), A^\top \lambda_1 -A^\top \lambda_2 \rangle \geq \nu\|x^*(\lambda_2) - x^*(\lambda_1)\|^2.
\end{equation*}
Noting that the right hand side
\begin{align*}
    \langle x^*(\lambda_2)-x^*(\lambda_1), A^\top \lambda_1 -A^\top \lambda_2 \rangle & \leq \|x^*(\lambda_2)-x^*(\lambda_1)\|\|A^\top (\lambda_1 -\lambda_2) \| \\
    & \leq \sigma_{\max}\|x^*(\lambda_2)-x^*(\lambda_1)\|\|\lambda_1 -\lambda_2 \|.
\end{align*}
Therefore,
\begin{equation*}
    \sigma_{\max}\|x^*(\lambda_2)-x^*(\lambda_1)\|\|\lambda_1 -\lambda_2 \| \geq \nu\|x^*(\lambda_2) - x^*(\lambda_1)\|^2,
\end{equation*}
which implies
\begin{equation*}
    \|x^*(\lambda_2) - x^*(\lambda_1)\| \leq \frac{\sigma_{\max}}{\nu}\|\lambda_1 - \lambda_2\|.
\end{equation*}
This implies that $x^*(\lambda)$ is $\frac{\sigma_{\max}}{\lambda}$-Lipschitz. By Danskin's Theorem, $\nabla h(\lambda) = -A x^*(\lambda)$. Then,
\begin{equation*}
    \|\nabla h(\lambda_1) - \nabla h(\lambda_2)\| = \|Ax^*(\lambda_1) - Ax^*(\lambda_2)\|\leq \sigma_{\max}\|x^*(\lambda_1) - x^*(\lambda_2)\| \leq \frac{\sigma_{\max}^2}{\nu}\|\lambda_1 - \lambda_2\|.
\end{equation*}
We conclude that $h$ is $\frac{\sigma_{\max}^2}{\nu}$-smooth. 

\vspace{2mm}
\noindent\textbf{Strong concavity:}  Since $f$ is $\ell$-smooth, for any $\lambda_1$ and $\lambda_2$,
\begin{align} \nonumber
    \langle x^*(\lambda_2)-x^*(\lambda_1), \nabla f(x^*(\lambda_2)) - \nabla f(x^*(\lambda_1))  \rangle  & \geq \frac{1}{\ell}\|\nabla f(x^*(\lambda_2)) - \nabla f(x^*(\lambda_1))\|^2 \\ \nonumber
    & = \frac{1}{\ell}\|A^\top \lambda_1 - A^\top \lambda_2\|^2 ,
\end{align}
where in the equality, we use $\nabla f(x^*(\lambda)) = -A^\top \lambda$ by the optimality of $x^*(\lambda)$. For $\lambda_1, \lambda_2 \in \spn(A)$, we have  $\lambda_1 - \lambda_2 \in \spn(A)$. Let the compact SVD of $A^\top$ be $A^\top = U\Sigma V^\top$. Then
\begin{align*}
    \|A^\top(\lambda_1 - \lambda_2) \|^2 & = \sum_{i = 1}^n (\lambda_1 - \lambda_2)_i^\top AA^\top (\lambda_1 - \lambda_2)_i \\
    & = \sum_{i = 1}^n \left(V^\top (\lambda_1 - \lambda_2)_i\right)^\top \Sigma^2 \left(V^\top (\lambda_1 - \lambda_2)_i\right) \\
    & \geq \sigma_{\min}^2\|V^\top (\lambda_1 - \lambda_2)\|^2 = \sigma_{\min}^2 \|\lambda_1 - \lambda_2\|^2,
\end{align*}
where in the last equality we use $\lambda_1 - \lambda_2 \in \spn(A)$. Combining with the last inequality,
\begin{equation*}
    \langle x^*(\lambda_2)-x^*(\lambda_1), \nabla f(x^*(\lambda_2)) - \nabla f(x^*(\lambda_1))  \rangle \geq \frac{\sigma_{\min}^2}{\ell}\|\lambda_1 - \lambda_2\|^2.
\end{equation*}
Combining with (\ref{eq:prop_optimality}),
\begin{equation*}
    \langle x^*(\lambda_2)-x^*(\lambda_1), A^\top \lambda_1 -A^\top \lambda_2 \rangle \geq \frac{\sigma_{\min}^2}{\ell}\|\lambda_1 - \lambda_2\|^2.
\end{equation*}
We further note that
\begin{equation*}
      \langle \nabla h(\lambda_2) - \nabla h(\lambda_1), \lambda_1-\lambda_2 \rangle = \langle Ax^*(\lambda_2) - Ax^*(\lambda_1), \lambda_1-\lambda_2 \rangle = \langle x^*(\lambda_2)-x^*(\lambda_1),  A^\top \lambda_1 -A^\top \lambda_2 \rangle .
\end{equation*}
Therefore,
\begin{equation*}
     \langle \nabla h(\lambda_2) - \nabla h(\lambda_1), \lambda_1-\lambda_2 \rangle \geq  \frac{\sigma_{\min}^2}{\ell}\|\lambda_1 - \lambda_2\|^2.
\end{equation*}
We conclude that $h$ is $\frac{\sigma_{\min}^2}{\ell}$-strongly concave.
\end{proof}

Now we discuss a version of Accelerated Gradient Descent (AGD) for $\min_{x\in \cX} f(x)$, where $\cX$ is convex and closed. In this algorithm, we can use inexact gradient $g(x)$ that is a biased gradient estimate for true gradient $\nabla f(x)$. 
\begin{align}  \label{eq:agd}
x_{t+1} & =\Pi_{\mathcal{X}}\left(y_t- \tau g\left(y_t\right)\right) \\ \nonumber
y_{t+1} & =x_{t+1}+\beta\left(x_{t+1}-x_t\right) .
\end{align}
The convergence guarantee for this algorithm is provided in \citep{devolder2013first, zhang2024optimal}. The following lemma is adopted from Lemma A.4 in \citep{zhang2024optimal}.

\begin{lemma}
\label{lemma:agd}
    Assume $f$ is $\mu$-strongly convex and $L$-smooth. Let $x^*$ be the unique minimizer of $f(x)$ and $\kappa=L/\mu$ be the condition number. Let $g(\cdot)$ be an inexact gradient oracle such that $\bE\|g(x) - \nabla f(x)\|^2 \leq \delta^2$, where the expectation is taken over the randomness in the gradient oracle. Starting from $y_0=x_0$, AGD described in \eqref{eq:agd} with $\tau = \frac{1}{2L}$ and $\beta = \frac{2-\sqrt{\mu / L}}{2+\sqrt{\mu / L}}$ satisfies
\begin{equation*}
\bE f\left(x_T\right)-f\left(x^*\right) \leq \exp \left(-\frac{T}{2 \sqrt{\kappa}}\right)\left(f\left(x_0\right)-f\left(x^*\right)+\frac{\mu}{4}\left\|x_0-x^*\right\|^2\right)+\sqrt{\kappa}\left(\frac{1}{L}+\frac{2}{\mu}\right) \delta^2 .
\end{equation*}
\end{lemma}

\begin{proof}
The lemma is identical to Lemma A.4 in \citep{zhang2024optimal} except that it accounts for the potential randomness in the inexact gradient oracle. Specifically, we replace the condition \(\|g(x) - \nabla f(x)\|^2 \leq \delta^2\) with \(\mathbb{E}\|g(x) - \nabla f(x)\|^2 \leq \delta^2\). Consequently, our guarantees for \(x_T\) hold in expectation as well. The  proof in \citep{zhang2024optimal} and \citep{devolder2013first} remains applicable.
\end{proof}

\subsection{Proofs for Acc-GA-MSGD}

\begin{lemma}
\label{lemma:dual-function-decentralized}
Under Assumption ~\ref{assum:global} and \ref{assum:sc}, the dual function $\Psi(\lambda) \triangleq \min_{X}\cL(X, \lambda)$ is $\frac{2M}{\mu}$-smooth, and $\frac{M(1-\sigma_2)}{L}$-strongly concave in $\spn(U)$.
\end{lemma}

\begin{proof}
Under Assumption~\ref{assume:W} for \(W\), with \(U = \sqrt{I - W}\), we know that the largest singular value of \(U\) is no larger than $\sqrt{2}$, and the smallest nonzero singular value of \(U\) is no smaller than $\sqrt{1-\sigma_2}$. We conclude by applying Proposition \ref{prop:duality} and Lemma \ref{prop:duality-decentralized}.
\end{proof}

\begin{theorem}
\label{thm:outer-loop-lambda-decentral}
Under Assumption~\ref{assum:global}, \ref{assume:W} and \ref{assum:sc}, we choose inner loop iterations $K$ large enough such that $\bE \|X^{t, K} - X^*(\lambda^t)\|^2 \leq \delta$, where $X^*(\lambda^t) = \argmin_{X} \cL(X, \lambda^t)$ is the optimal primal variable corresponding to $\lambda^t$. With stepsize  $\tau_2 = \frac{\mu}{4M}$, $\beta = \frac{2\sqrt{2L} - \sqrt{(1-\sigma_2)\mu}}{2\sqrt{2L} + \sqrt{(1-\sigma_2)\mu}}$ and $\lambda^0 \in \spn(U)$, the last iterate of Algorithm~\ref{alg:acc-multi-agda}, $\lambda^T$, satisfies:
\begin{equation*}
\bE\|\lambda^T - \lambda^*\|^2 \leq \exp \left(-\frac{T\sqrt{\mu(1-\sigma_2)}}{2 \sqrt{2L}}\right)\frac{1}{2}\left\|\lambda^0-\lambda^*\right\|^2 + \frac{8L^{\frac{5}{2}}}{M^2\sqrt{\mu}(1-\sigma_2)^{\frac{5}{2}}} \delta.
\end{equation*}
where $\lambda^*$ is the optimal dual solution to $\cL( X, \lambda)$ in $\spn(U)$.

\end{theorem}

\begin{proof}
Because $ \cL^s(X, \lambda)$ is strongly convex in $X$, by Danskin's Theorem, 
$$
\nabla \Psi(\lambda) = UX^*(\lambda).
$$
Then we can bound
\begin{align*}
    \bE \|\nabla \Phi(\lambda^t) - UX^{t, K}\|^2 & = \bE \|UX^*(\lambda) - UX^{t, K}\|^2 \\
    & \leq 2\bE\|X^*(\lambda) - X^{t, K}\|^2 \leq 2\delta.
\end{align*}
Algorithm~\ref{alg:acc-multi-agda} can be considered equivalent to running AGD (\ref{eq:agd}) on -$\Psi(\lambda)$ on the subspace $\spn(U)$ with an inexact gradient oracle. Note that by the update rule of $\lambda$ (\ref{eq:lambda-update}), the sequence $\{\lambda^t \}$ stays in $\spn(U)$ as we pick the initial point $\lambda^0 = 0 \in \spn(U)$. Applying Lemma~\ref{lemma:agd} and Lemma~\ref{lemma:dual-function-decentralized}, 
\begin{equation*}
\Psi(\lambda^*) - \bE \Psi\left(\lambda^T\right) \leq \exp \left(-\frac{T\sqrt{\mu(1-\sigma_2)}}{2 \sqrt{2L}}\right)\frac{M(1-\sigma_2)}{4L}\left\|\lambda^0-\lambda^*\right\|^2 + \sqrt{\frac{2L}{\mu(1-\sigma_2)}}\left(\frac{\mu}{2M}+\frac{2L}{M(1-\sigma_2)}\right) \delta .
\end{equation*}
Since $\Psi$ is $\frac{M(1-\sigma_2)}{L}$-strongly concave, 
\begin{align*}
\bE\|\lambda^T - \lambda^*\|^2 & \leq \exp \left(-\frac{T\sqrt{\mu(1-\sigma_2)}}{2 \sqrt{2L}}\right)\frac{1}{2}\left\|\lambda^0-\lambda^*\right\|^2 + \sqrt{\frac{2L}{\mu(1-\sigma_2)}}\frac{2L}{M(1-\sigma_2)}\left(\frac{\mu}{2M}+\frac{2L}{M(1-\sigma_2)}\right) \delta \\
& \leq \exp \left(-\frac{T\sqrt{\mu(1-\sigma_2)}}{2 \sqrt{2L}}\right)\frac{1}{2}\left\|\lambda^0-\lambda^*\right\|^2 + \frac{8L^{\frac{5}{2}}}{M^2\sqrt{\mu}(1-\sigma_2)^{\frac{5}{2}}} \delta.
\end{align*}

\end{proof}

\paragraph{Proof for  Theorem \ref{thm:outer-loop-decentral}}
\mbox{}

\begin{proof}
Note that 
\begin{align*}
     \fakeeq \bE \|X^{T, K} - X^*\|^2 & \leq 2\bE \left[\|X^{T, K}-X^*(\lambda^T)\|^2 + \|X^*(\lambda^T) - X^*\|^2\right]\\
    & \leq 2\delta + 2\bE\|X^*(\lambda^T) - X^*(\lambda^*)\|^2.
\end{align*}
Since $ \cL(X, \lambda)$ is $\frac{\mu}{M}$-strongly convex and the largest singular value of $U$ is upper bounded by $\sqrt{2}$, by Proposition \ref{prop:duality}, we have
$$
\|X^*(\lambda^T) - X^*(\lambda^*)\|^2 \leq \frac{2M^2}{\mu^2}\|\lambda^T - \lambda^*\|^2 .
$$
Therefore, 
\begin{equation*}
    \bE \|X^{T, K} - X^*\|^2 \leq 2\delta + \frac{4M^2}{\mu^2}\bE \|\lambda^T - \lambda^*\|^2.
\end{equation*}
By Jensen's inequality,
$$
\bE \|\bar x^T -x^*_s\|^2 \leq \frac{1}{M}\sum_{m=1}^M\bE\|x_m^{T, K} - x^*\|^2 \leq \frac{2\delta}{M} + \frac{4M}{\mu^2}\bE \|\lambda^T - \lambda^*\|^2.
$$
Because $F(x)$ is $L$-smooth, 
\begin{align} \nonumber
    \bE F(\bar x^T) - \min_x F(x) & \leq \frac{L}{2}\bE \|\bar x^T - x^*\|^2 \\ \nonumber
    & \leq \frac{L\delta}{M} + \frac{2ML}{\mu^2}\bE \|\lambda^T - \lambda^*\|^2  \\ \nonumber
    &\leq \frac{L\delta}{M} + \frac{2ML}{\mu^2} \left[ \exp \left(-\frac{T\sqrt{\mu(1-\sigma_2)}}{2 \sqrt{2L}}\right)\frac{1}{2}\left\|\lambda^0-\lambda^*\right\|^2 + \frac{8L^{\frac{5}{2}}}{M^2\sqrt{\mu}(1-\sigma_2)^{\frac{5}{2}}} \delta \right] \\
    & \leq \frac{ML}{\mu^2} \exp \left(-\frac{T\sqrt{\mu(1-\sigma_2)}}{2 \sqrt{2L}}\right)\left\|\lambda^0-\lambda^*\right\|^2  + \frac{17L^{\frac{7}{2}}}{M\mu^{\frac{5}{2}} (1-\sigma_2)^\frac{5}{2}} \delta.
\end{align}
where in the their inequality we use Theorem~\ref{thm:outer-loop-lambda-decentral}. By choosing $\lambda^0 = 0$, 
\begin{align}
\label{eq:initial-lambda-bd-decen}
    \|\lambda^0 - \lambda^*\|^2 = \| \lambda^*\|^2 \leq \frac{\sum_{m=1}^M\|\nabla F_m(x^*)\|^2}{1-\sigma_2} \leq  \frac{L^2\|X_0 - X^*\|^2}{1-\sigma_2},
\end{align}
where in the second inequality we use Theorem 3 in \citep{lan2020communication}, and in the last inequality we use the smoothness of each $F_m$.
Therefore, 
\begin{align*}
    \bE F(\bar x^T) - \min_x F(x) \leq \frac{ML}{\mu^2} \exp \left(-\frac{T\sqrt{\mu(1-\sigma_2)}}{2 \sqrt{2L}}\right) \frac{L^2\|X_0 - X^*\|^2}{1-\sigma_2}  + \frac{17L^{\frac{7}{2}}}{M\mu^{\frac{5}{2}} (1-\sigma_2)^\frac{5}{2}} \delta,
\end{align*}
Now we evaluate the consensus
\begin{align*}
    \frac{1}{M}\sum_{m=1}^M\|X_m^{T, K} - \bar x^T\|^2 & \leq \frac{2}{1-\sigma_2}\|UX^{T, K}\|^2 \\
    & = \frac{2}{(1-\sigma_2)\tau_2^2}\|\lambda^{T+1} - \widetilde \lambda^T\|^2 \\
    & \leq \frac{2}{(1-\sigma_2)\tau_2^2}\|\lambda^{T+1} - [\lambda^T + \beta(\lambda^T - \lambda^{T-1})]\|^2 \\
    & \leq \frac{2}{(1-\sigma_2)\tau_2^2} \left[2\|\lambda^{T+1} -\lambda^T\|^2 + 2\beta^2\|\lambda^T - \lambda^{T-1}\|^2\right] \\
    & \leq \frac{8}{(1-\sigma_2)\tau_2^2}\left[\|\lambda^{T+1} - \lambda^*\|^2 + 2\|\lambda^{T} - \lambda^*\|^2 +\|\lambda^{T-1} - \lambda^*\|^2 \right]
\end{align*}
where in the first inequality we use Lemma 3 in \citep{li2020revisiting}. Taking the expectation and applying Theorem \ref{thm:outer-loop-lambda-decentral},
\begin{align*}
     \frac{1}{M}\sum_{m=1}^M \bE\|X_m^{T, K} - \bar x^T\|^2 \leq \frac{256M^2}{(1-\sigma_2)\mu^2}\exp \left(-\frac{T\sqrt{\mu(1-\sigma_2)}}{2 \sqrt{2L}}\right)\left\|\lambda^0-\lambda^*\right\|^2 + \frac{4096L^{\frac{5}{2}}}{\mu^{\frac{5}{2}}(1-\sigma_2)^{\frac{7}{2}}} \delta.
\end{align*}
We conclude by combining with (\ref{eq:initial-lambda-bd-decen}).

\end{proof}

\paragraph{Proof for Lemma \ref{thm:inner-loop-decentral}}
\mbox{}

\begin{proof}
    $\cL( X, \lambda)$ is $\mu/M$-strongly convex and $L/M$-smooth in $X$. Also, we have access to a stochastic gradient estimate for $\nabla_X  \cL( X, \lambda)$ with variance $\frac{\sigma^2}{M}$.
By the convergence rate of SGD on strongly convex and smooth functions (e.g., Theorem 5 of \citep{stich2019unified}), to achieve $\bE\|X^{t, K} - X^*(\lambda^t)\|^2 \leq \delta = \frac{M\mu^{\frac{5}{2}} (1-\sigma_2)^\frac{5}{2}\epsilon}{34L^{\frac{7}{2}}}$,
\begin{equation} \label{eq:inner-loop-bd-decen}
    K = \cO\left(\kappa\log\left(\frac{L^{\frac{9}{2}}\bE\|X^{t,0} -  X^*(\lambda^t)\|^2}{\mu^{\frac{7}{2}}M(1-\sigma_2)^{\frac{5}{2}}\epsilon} \right) + \frac{L^{\frac{7}{2}}\sigma^2}{\mu^{\frac{9}{2}}(1-\sigma_2)^{\frac{5}{2}}\epsilon} \right).
\end{equation}
Now we bound $\bE\|X^{t,0} -  X^*(\lambda^t)\|^2$. When $t \geq 1$, 
\begin{align} \nonumber
    \bE\|X^{t,0} -  X^*(\lambda^t)\|^2 & \leq 2\bE\|X^{t,0} -  X^*(\lambda^{t-1})\|^2 + 2\bE\|X^*(\lambda^{t-1}) -  X^*(\lambda^t)\|^2 \\ \nonumber
    & \leq 2\bE\|X^{t,0} -  X^*(\lambda^{t-1})\|^2 + \frac{4M^2}{\mu^2} \bE\|\lambda^{t-1}- \lambda^t\|^2 \\  \label{eq:initial-x0-de}
    & \leq \frac{M\mu^{\frac{5}{2}} (1-\sigma_2)^\frac{5}{2}\epsilon}{17L^{\frac{7}{2}}} + \frac{4M^2}{\mu^2} \left[ \bE\|\lambda^{t-1}- \lambda^*\|^2 + \bE\|\lambda^{t}- \lambda^*\|^2  \right],
\end{align}
where we use Proposition~\ref{prop:duality} in the second inequality.  According to Theorem \ref{thm:outer-loop-lambda-decentral}, for all $t \geq 1$,
\begin{align*}
\bE\|\lambda^T - \lambda^*\|^2 & \leq \exp \left(-\frac{T\sqrt{\mu(1-\sigma_2)}}{2 \sqrt{2L}}\right)\frac{1}{2}\left\|\lambda^0-\lambda^*\right\|^2 + \frac{8L^{\frac{5}{2}}}{M^2\sqrt{\mu}(1-\sigma_2)^{\frac{5}{2}}} \delta\\
& \leq \frac{1}{2}\left\|\lambda^0-\lambda^*\right\|^2 + \cO\left( \frac{\mu^2\epsilon}{LM}\right) \\
& \leq  \frac{L^2\|X_0 - X^*\|^2}{2(1-\sigma_2)} + \cO\left( \frac{\mu^2\epsilon}{LM}\right).
\end{align*}
where in the last inequality we use (\ref{eq:initial-lambda-bd-decen}). Therefore, 
\begin{equation*}
     \bE\|X^{t,0} -  X^*(\lambda^t)\|^2 \leq \cO\left( \frac{M\mu^{\frac{5}{2}} (1-\sigma_2)^\frac{5}{2}\epsilon}{L^{\frac{7}{2}}} + \frac{M^2L^3}{\mu^3}\|X^0 - X^*\|^2 \right).
\end{equation*}
Combining with (\ref{eq:inner-loop-bd-decen}), 
\begin{equation*}
     K = \cO\left(\kappa\log\left( \frac{L^{\frac{15}{2}}M\|X^0 - X^*\|^2}{\mu^{\frac{13}{2}}(1-\sigma_2)^{\frac{5}{2}}\epsilon} + \frac{L}{\mu} \right) + \frac{L^{\frac{7}{2}}\sigma^2}{\mu^{\frac{9}{2}}(1-\sigma_2)^{\frac{5}{2}}\epsilon} \right).
\end{equation*}
When $t = 0$, 
\begin{align*}
    \|X^{0,0} -  X^*(\lambda^0)\|^2 & \leq 2 \|X^{0,0} -  X^*\|^2 + 2\|X^* -  X^*(\lambda^0)\|^2 \\
    & \leq 2 \|X^0 - X^*\|^2 + \frac{4M^2}{\mu^2}\|\lambda^0 - \lambda^*\|^2 \\
    & \leq 2 \|X^0 - X^*\|^2 +\frac{4L^2M^2\|X^0 - X^*\|^2}{\mu^2(1-\sigma_2)}.
\end{align*}
Combining with (\ref{eq:inner-loop-bd-decen}), 
\begin{equation*}
     K = \cO\left(\kappa\log\left(\frac{L^{\frac{13}{2}}M\|X^{0} -  X^*\|^2}{\mu^{\frac{11}{2}}(1-\sigma_2)^{\frac{7}{2}}\epsilon} \right) + \frac{L^{\frac{7}{2}}\sigma^2}{\mu^{\frac{9}{2}}(1-\sigma_2)^{\frac{5}{2}}\epsilon} \right).
\end{equation*}

\end{proof}

\subsection{Catalyst Acceleration for Acc-GA-MSGD}
\label{subsec:cata-decentral}

\begin{algorithm}[t] 
    \caption{Decentralized Catalyst}
    \begin{algorithmic}[1] 
        \State \textbf{Initialization}: initial point $x_0$. In the strongly convex setting, $q = \frac{\mu}{\mu + 2L}$ and $\alpha_1 = \sqrt{q}$; in the convex and nonconvex settings, $q = 0$ and $\alpha_1 = 1$. Initialize $y^0 = x^0$.
        \For{$s = 1,2,..., S$}
            \State Find an inexact solution $(x_1^s, x_2^s, \dots, x_M^s)$ to the following decentralized problem
            \vspace{-1mm}
            \begin{equation} \label{cata:subproblem-decentral}
            \min_{x} F^s(x) = \frac{1}{M}\sum_{m=1}^M F_m^s \text{  with  } F_m^s(x) = F_m(x) + L\|x - y_m^s\|^2 .
            \vspace{-1mm}
            \end{equation} 
           \quad \ \ such that $\bE F^s(\bar x^s) - \min_x F^s(x) \leq \epsilon^s$ with $\bar x^s = \frac{1}{M}\sum_{m=1}^M x_m^s$. \vspace{1mm}
            \State Update momentum parameters: in the strongly convex and convex settings, $\alpha_s^2=\left(1-\alpha_s\right) \alpha_{s-1}^2+q \alpha_s$ and $\beta_s=\frac{\alpha_{s-1}\left(1-\alpha_{s-1}\right)}{\alpha_{s-1}^2+\alpha_s}$; in the nonconvex setting, $\beta_s = 0$.
            \State Compute the new prox center: $y_m^{s+1} = x_m^s + \beta_s(x_m^s - x_m^{s-1})$ for $m=1, 2,\dots, M$
        \EndFor
    \end{algorithmic} 
\label{alg:APPA-decentral}
\end{algorithm}

\citet{li2020revisiting} adapted Catalyst to the decentralized setup, which is described in Algorithm~\ref{alg:APPA-decentral}. In particular, in each iteration it approximately solved a decentralized problem with a decentralized algorithm:
\begin{equation*}
\min_{x} F^s(x) = \frac{1}{M}\sum_{m=1}^M F_m^s \text{  with  } F_m^s(x) = F_m(x) + L\|x - y_m^s\|^2 .
\end{equation*}
Note that with $\bar y^s = \frac{1}{M}\sum_{m=1}^M y_m^s$,
\begin{align*}
    F^s(x) & = \frac{1}{M}\sum_{m=1}^M F_m(x) + \frac{1}{M}\sum_{m=1}^M  L\|x - y_m^s\|^2 \\
    & = \frac{1}{M}\sum_{m=1}^M F_m(x) + L\|x - \bar y^s\|^2 - L\|\bar y^s\|^2 + \frac{L}{M}\sum_{m=1}^M\|y_m^s\|^2.
\end{align*}
The subproblem (\ref{cata:subproblem-decentral}) is equivalent to finding an approximate solution $\bar x^s$ to 
$$
\min_x G^s(x) = \frac{1}{M}\sum_{m=1}^M F_m(x) + L\|x - \bar y^s\|^2.
$$
Therefore, the criterion $\bE F^s(\bar x^s) - \min_x F^s(x) \leq \epsilon^s$ in Algorithm~\ref{alg:APPA-decentral} is equivalent to $\bE G^s(\bar x^s) - \min_x G^s(x) \leq \epsilon^s$. Moreover, we observe that 
$$
\bar y^{s+1} = \bar x^s + \beta_s (\bar x^s - \bar x^{s-1}).
$$
Consequently, Algorithm~\ref{alg:APPA-decentral} can be considered as applying Algorithm~\ref{alg:APPA} to $\bar x$. Theorem~\ref{thm:catalyst} can be directly used to provide a convergence guarantee for $\bar x^S$.

\paragraph{Proof for Corollary \ref{coro:catalyst-acc}}
\mbox{}

\begin{proof}
Based on our discussion above, the same proof reasoning of Corollary~\ref{coro:catalyst-GA-MSGD} can be adopted here. 
The subproblem (\ref{cata:subproblem-decentral}) has condition number being $\cO(1)$. According to Corollary~\ref{coro:total-complexity-decentral}, to find a solution $(x_1^s, x_2^s, \dots, x_M^s)$ such that $F^s(\bar x) - \min_x F^s(x) \leq \epsilon^s$ and $\frac{1}{M}\sum_{m=1}^M \bE\|x^s_m - \bar x^s\|^2 \leq \cO(M\epsilon^s/L)$ with $\bar x^s\triangleq \frac{1}{M}\sum_{m=1}^M x^s_m$, the total number of communication rounds in Algorithm \ref{alg:acc-multi-agda} is 
 \begin{equation*}
       T = \cO \left( \sqrt{\frac{1}{1-\sigma_2}} \log\left( \frac{ LM\|X_s^0 - X_s^*\|^2}{\epsilon^s}\right)\right).
   \end{equation*}
The total number of local gradients (samples) used by each client is 
\begin{equation*}
       TK = \cO\left(\frac{1}{\sqrt{1-\sigma_2}}\log^2\left( \frac{L M\|X_s^0 - X_s^*\|^2}{(1-\sigma_2)\epsilon^s} + 1 \right) + \frac{\sigma^2}{\mu(1-\sigma_2)^{3}\epsilon^s}\log\left( \frac{LM\|X_s^0 - X_s^*\|^2}{\epsilon^s}\right) \right),
   \end{equation*}
   where $X_s^0$ denotes the initial point of Acc-GA-MSGD in solving $s$-th subproblem, $X_s^* = (x_s^* \dots x_s^*)^\top$ with \(x_s^*\) being the optimal solution for $s$-th subproblem, and $\epsilon^s$ is the target accuracy.

\paragraph{Strongly Convex.} By Theorem~\ref{thm:catalyst}, the number of Catalyst iterations to achieve $\bE F(\bar x^S) - F^* \leq \epsilon$ is 
$$
S = \widetilde\cO(\sqrt{\kappa}).
$$
As in the proof of Corollary~\ref{coro:catalyst-GA-MSGD}, we have
\begin{equation*}
    \epsilon^S  = \Theta\left( \frac{\epsilon}{\kappa}\right).
\end{equation*}
When running Acc-GA-MSGD to guarantee $\bE F^S(\bar x^S) - \min_x F^S(x) \leq \epsilon^S$, it also guarantees $\frac{1}{M}\sum_{m=1}^M \bE\|x^S_m - \bar x^S\|^2 \leq \cO(M\epsilon^S/L) = \cO(M\epsilon/(\kappa L))$. Moreover, for all $s$, Acc-GA-MSGD guarantees that $\frac{1}{M}\sum_{m=1}^M \bE\|x^s_m - \bar x^s\|^2 \leq \cO(M\epsilon^s/L)$ is upper bounded. We pick $x_{s, m}^0  = x_m^{s-1} + \frac{2L}{2L + \mu}(y_m^s - y_m^{s-1})$ for all $m$ and $X_s^0 = (x_{s, 1}^0 \cdots x_{s, M}^0 )^\top$ as the initial point for $s$-th round. Recall our discussion at the beginning of this subsection that Algorithm~\ref{alg:APPA-decentral} can be considered as applying Algorithm~\ref{alg:APPA} to $\bar x$. Note that \(\bar x_s^0 = \frac{1}{M}\sum_{m=1}^M x_{s,m} = \bar x^{s-1} + \frac{2L}{2L+\mu} (\bar y^s - \bar y^{s-1})\). According to Proposition 12 in \citep{lin2018catalyst}, $\bE \|x_s^0 - x_s^*\|^2$ is bounded. Because $\frac{1}{M}\sum_{m=1}^M \bE\|x^s_m - \bar x^s\|^2$ is upper bounded, with the update rule for $y_m^s$ in Algorithm~\ref{alg:APPA-decentral}, it is easy to show that $\frac{1}{M}\sum_{m=1}^M \bE\|y^s_m - \bar y^s\|^2$ is also bounded. Therefore, by definition of $x_{s,m}^0$, we have $\frac{1}{M}\sum_{m=1}^M \bE\|x^0_{s,m} - \bar x_0^s\|^2$ is bounded for all $s$. As
\begin{align*}
    \bE\|X_s^0 - X_s^*\|^2 = \bE\sum_{m=1}^M \| x_{s,m}^0 - x_s^*\|^2 \leq 2\bE\sum_{m=1}^M\|x_{s, m}^0 - \bar x_s\|^2 + 2 \bE \sum_{m=1}^M\|\bar x_s - x_s^*\|^2,
\end{align*}
we conclude that $\bE\|X_s^0 - X_s^*\|^2$ is upper bounded for all $s$. Together with $\epsilon^s \geq \Theta(\epsilon/\kappa)$ as demonstrated in the proof of Corollary~\ref{coro:catalyst-GA-MSGD}, we have $T_s = \widetilde \cO\left(\frac{1}{\sqrt{1-\sigma_2}}\right)$ and $T_sK_s = \widetilde \cO\left(\frac{1}{\sqrt{1-\sigma_2}} + \frac{\kappa\sigma^2}{L\epsilon}(1-\sigma_2)^3 \right)$. The total communication rounds is $\sum_{s}T_s = \widetilde \cO\left(\frac{\sqrt \kappa}{\sqrt{1-\sigma_2}}\right)$, and total sample complexity for each client is $\sum_{s}T_sK_s = \widetilde \cO\left(\frac{\sqrt \kappa}{\sqrt{1- \sigma_2}} + \frac{\kappa^{1.5}\sigma^2}{L(1-\sigma_2)^3\epsilon}\right) = \widetilde \cO\left(\frac{\sqrt \kappa}{\sqrt{1- \sigma_2}} + \frac{\sqrt \kappa\sigma^2}{\mu(1-\sigma_2)^3\epsilon}\right)$.

\paragraph{Convex. }  By Theorem~\ref{thm:catalyst}, the number of Catalyst iterations to achieve $\bE F(\bar x^S) - F^* \leq \epsilon$ is 
$$
S = \widetilde\cO\left(\sqrt{\frac{L\|x_0 - x^*\|^2}{\gamma^2\epsilon}} \right).
$$
As in the proof of Corollary~\ref{coro:catalyst-GA-MSGD}, we have 
\begin{equation*}
    \epsilon^S  = \Theta\left( \frac{\gamma^{4+\gamma}\epsilon^{2+\frac{\gamma}{2}}}{\Delta^{1+\frac{\gamma}{2}}}\right).
\end{equation*}
When running Acc-GA-MSGD to guarantee $\bE F^S(\bar x^S) - \min_x F^S(x) \leq \epsilon^S$, it also guarantees $\frac{1}{M}\sum_{m=1}^M \bE\|x^S_m - \bar x^S\|^2 \leq \cO(M\epsilon^S/L) = \cO\left( \frac{M \gamma^{4+\gamma}\epsilon^{2+\frac{\gamma}{2}}}{L \Delta^{1+\frac{\gamma}{2}}}\right)$.
We pick $x_{s, m}^0  = x_m^{s-1} + \frac{2L}{2L + \mu}(y_m^s - y_m^{s-1})$ for all $m$ and $X_s^0 = (x_{s, 1}^0 \cdots x_{s, M}^0 )^\top$ as the initial point for $s$-th round. It can be shown, similar to the strongly convex case, that $\bE\|X_s^0 - X_s^*\|^2$ is upper bounded for all $s$. Together with $\epsilon^s \geq \Theta\left( \frac{\gamma^{4+\gamma}\epsilon^{2+\frac{\gamma}{2}}}{\Delta^{1+\frac{\gamma}{2}}}\right)$ as demonstrated in the proof of Corollary~\ref{coro:catalyst-GA-MSGD},
 we have $T_s = \widetilde \cO\left(\frac{1}{\sqrt{1-\sigma_2}}\right)$ and $T_sK_s = \widetilde \cO\left(\frac{1}{\sqrt{1-\sigma_2}} + \frac{\Delta^{1+\frac{\gamma}{2}}\sigma^2}{(1-\sigma_2)^3L\gamma^{4+\gamma}\epsilon^{2+\frac{\gamma}{2}}} \right)$. The total communication rounds is $\sum_{s}T_s = \widetilde \cO\left(\sqrt{\frac{L\|x_0 - x^*\|^2}{\epsilon}}\right)$, and total sample complexity for each client is $\sum_{s}T_sK_s = \widetilde \cO\left(\sqrt{\frac{L\|x_0 - x^*\|^2}{(1-\sigma_2)\epsilon}} + \frac{\Delta^{1+\frac{\gamma}{2}}\sigma^2\|x_0 - x^*\|^2}{(1-\sigma_2)^3\sqrt{L}\gamma^{4+\gamma}\epsilon^{2.5+\frac{\gamma}{2}}}\right)$.

\paragraph{Nonconvex. }  By Theorem~\ref{thm:catalyst}, the number of Catalyst iterations to achieve $\bE\|\nabla F(\bar x)\|^2 \leq \epsilon^2$ is 
$$
S = \widetilde\cO\left(\frac{L\Delta}{\epsilon^2}\right).
$$
As in the proof of Corollary~\ref{coro:catalyst-GA-MSGD}, we have for all $s$
\begin{equation*}
    \epsilon^s = \frac{\epsilon^2}{32L}.
\end{equation*}
When running Acc-GA-MSGD to guarantee $\bE F^s(\bar x^s) - \min_x F^s(x) \leq \epsilon^s$, it also guarantees $\frac{1}{M}\sum_{m=1}^M \bE\|x^s_m - \bar x^s\|^2 \leq \cO(M\epsilon^s/L) = \cO\left( \frac{M\epsilon^2}{L^2}\right)$ for all $s$. We pick $x_{s, m}^0  = x_m^{s-1}$ for all $m$ and $X_s^0 = (x_{s, 1}^0 \cdots x_{s, M}^0 )^\top$ as the initial point for $s$-th round. We have $\frac{1}{M}\sum_{m=1}^M \bE\|x^0_{s,m} - \bar x_0^s\|^2$ is bounded for all $s$. This further implies  that $\bE\|X_s^0 - X_s^*\|^2$ is upper bounded.
Together with \(\epsilon^s = \epsilon^2/(32L) \),  we have $T_s = \widetilde \cO\left(\frac{1}{\sqrt{1-\sigma}}\right)$ and $T_sK_s = \widetilde \cO\left(\frac{1}{\sqrt{1-\sigma}} + \frac{\sigma^2}{(1-\sigma_2)^3\epsilon^2}\right)$. The total communication rounds is $\sum_{s}T_s = \widetilde \cO\left(\frac{L\Delta}{\sqrt{1-\sigma_2}\epsilon^2}\right)$, and total sample complexity for each client is $\sum_{s}T_sK_s = \widetilde \cO\left(\frac{L\Delta}{\sqrt{1-\sigma_2}\epsilon^2} +  \frac{L\Delta\sigma^2}{(1-\sigma_2)^3\epsilon^4}\right)$.

\end{proof}

\subsection{Better Communication Complexity for LED}
\label{subsec:LED}

 Local Exact Diffusion (LED), presented in Algorithm~\ref{alg:led}, was introduced by \citet{alghunaim2024local}. We observe that it is equivalent to applying Gradient Ascent-Multiple Stochastic Gradient Descent (GA-MSGD) to Lagrangian (\ref{eq:lagrangian-decen}):
\begin{align} \nonumber
    & X^{t, k+1} = X^{t, k} - \tau_1^k\left[\nabla H(X^{t, k}; \xi) + U\lambda^t \right], \quad k = 1, 2, \dots, K \\  
    &\lambda^{t+1} = \lambda^t + \tau_2 UX^{t, K}.
\end{align}
Define $\zeta = U\lambda$, and pre-multiply the update rule of $\lambda$ by $U$. Then it can be written as
\begin{align*}
    & X^{t, k+1} = X^{t, k} - \tau_1^k\left[ \nabla H(X^{t, k}; \xi) +  \zeta^t \right], \quad k = 1, 2, \dots, K\\
    &\zeta^{t+1} = \zeta^t + \tau_2 (I-W)X^{t, K},
\end{align*}
which is precisely the update rule (5) in \citep{alghunaim2024local}. When sufficient local steps are taken or the gradient is non-noisy, the original proof of LED establishes that it achieves communication complexity of $\widetilde\cO(\frac{\kappa^2}{1-\sigma_2})$ in the strongly convex setting (see Section C of \citep{alghunaim2024local}). In the following, we improve it to $\widetilde\cO(\frac{\kappa}{1-\sigma_2})$ based on our observation in Lemma~\ref{lemma:dual-function-decentralized} that the dual function $\Psi(\lambda) \triangleq \min_{X}\cL(X, \lambda)$ is $\frac{2M}{\mu}$-smooth, and $\frac{M(1-\sigma_2)}{L}$-strongly concave in $\spn(U)$.

\begin{theorem}
Under Assumption~\ref{assum:global}, \ref{assume:W} and \ref{assum:sc}, we choose inner loop iterations $K$ large enough such that   $\bE \|X^{t, K} - X^*(\lambda)\|^2 \leq \delta$,  where $X^*(\lambda^t) = \argmin_{X} \cL(X, \lambda^t)$ is the optimal primal variable corresponding to $\lambda^t$.. With stepsize  $\tau_2 = \frac{\mu}{2M}$, the last iterate of LED satisfies:
\begin{equation*}
    \bE \|\lambda^T - \lambda^*\|^2 \leq \left(1- \frac{(1-\sigma_2 )\mu}{4L}\right)^T \|\lambda^0 - \lambda^*\|^2  + \frac{16L^2}{(1-\sigma_2)^2M^2}\delta.
\end{equation*}
where $\lambda^*$ is the optimal dual solution to $\cL(X, \lambda)$.

\end{theorem}

\begin{proof}
Because $ \cL(X, \lambda)$ is strongly convex in $X$, by Danskin's Theorem, 
$$
\nabla \Psi(\lambda) = UX^*(\lambda).
$$
Define a helper point $\tilde \lambda^{t+1}$ that is from doing one step of gradient ascent from $\lambda^t$ using the exact gradient of $\Phi$:
$$\Tilde{\lambda}^{t+1} = \lambda^t + \tau_2 \nabla \Phi(\lambda^t) = \lambda_t + \tau_2 U X^*(\lambda^t). $$
Since we pick $\lambda^0 \in \spn(U)$, then $\{\lambda^t\}_t$ will stay in $\spn(U)$ due to the update rule of $\lambda^t$. Also, Lemma 3.1 of \citep{shi2015extra} shows that there exists $\lambda^* \in \spn(U)$. By Lemma~\ref{lemma:dual-function-decentralized}, $\Psi(\lambda)$ is $\frac{2M}{\mu}$-smooth, and $\frac{M(1-\sigma_2 )}{L}$-strongly concave in $\spn(U)$. Now we can follow the classic analysis of gradient for strongly concave function:
\begin{align*}
    \|\lambda^{t+1} - \lambda^*\|^2 & = \| \lambda^t + \tau_2 \nabla \Phi(\lambda^t) \|^2 \\
    & = \|\lambda^t - \lambda^*\|^2 + 2\tau_2 \langle \nabla \Phi(\lambda), \lambda^t- \lambda^*\rangle + \tau_2^2\|\nabla \Phi(\lambda)\|^2 \\
    &\leq \|\lambda^t - \lambda^*\|^2 - 2\tau_2\left(\frac{M(1-\sigma_2) }{2L}\|\lambda^t - \lambda^*\|^2 + \Phi(\lambda^t) - \Phi(\lambda^*) \right) + \frac{4M\tau_2^2}{\mu}\left( \Phi(\lambda^t) - \Phi(\lambda^*)\right),
\end{align*}
where in the inequality we use Lemma \ref{prop:duality-decentralized} and a equivalent definition of strong concavity. With $\tau_2 = \frac{\mu}{2M}$, 
\begin{equation}
    \|\Tilde{\lambda}^{t+1} - \lambda^*\|^2 \leq \left(1- \frac{(1-\sigma_2 )\mu}{2L}\right)\|\lambda^t - \lambda^*\|^2.
\end{equation}
By Young's Inequality, 
\begin{align*}
    \|\lambda^{t+1} - \lambda^*\|^2 & \leq \left(1+\frac{(1-\sigma_2 )\mu}{4L - (1-\sigma_2)\mu }\right)\|\Tilde{\lambda}^{t+1} - \lambda^*\|^2 + \left(1+\frac{4L - (1-\sigma_2)\mu }{(1-\sigma_2 )\mu}\right) \|\Tilde{\lambda}^{t+1} - \lambda^{t+1}\|^2 \\
    & \leq \left(1- \frac{(1-\sigma_2 )\mu}{4L}\right)\|\lambda^t - \lambda^*\|^2 + \frac{4L }{(1-\sigma_2 )\mu}\|\Tilde{\lambda}^{t+1} - \lambda^{t+1}\|^2 \\
    & =  \left(1- \frac{(1-\sigma_2 )\mu}{4L}\right)\|\lambda^t - \lambda^*\|^2 + \frac{4L\tau_2^2}{(1-\sigma_2 )\mu} \|UX^*(\lambda^t) - UX^{t, K} \|^2 \\
    & =  \left(1- \frac{(1-\sigma_2 )\mu}{4L}\right)\|\lambda^t - \lambda^*\|^2 + \frac{4L\mu}{(1-\sigma_2 )M^2} \|X^*(\lambda^t) - X^{t, K} \|^2.
\end{align*}
Taking expectation, 
\begin{align*}
    \bE \|\lambda^{t+1} - \lambda^*\|^2 &\leq \left(1- \frac{(1-\sigma_2 )\mu}{4L}\right)\bE\|\lambda^t - \lambda^*\|^2 + \frac{4L\mu}{(1-\sigma_2 )M^2} \bE\|X^*(\lambda^t) - X^{t, K} \|^2 \\
   & \leq \left(1- \frac{(1-\sigma_2 )\mu}{4L}\right)\bE\|\lambda^t - \lambda^*\|^2  + \frac{4L\mu}{(1-\sigma_2 )M^2}  \delta.
\end{align*}
Recursing this inequality,
\begin{equation*}
    \bE \|\lambda^T - \lambda^*\|^2 \leq \left(1- \frac{(1-\sigma_2 )\mu}{4L}\right)^T \|\lambda^0 - \lambda^*\|^2  + \frac{16L^2}{(1-\sigma_2)^2M^2}\delta.
\end{equation*}
\end{proof}

\begin{algorithm}[t]  
    \caption{LED}
    \begin{algorithmic}[1] 
      \State \textbf{Input:} initial point $x_1^0, x_2^0, \dots, x_M^0$, 
      and $\lambda^0$.
        \For{$t = 0,1,2,\dots, T$}
            \For{$k = 0,1,2,..., K-1$}
                 \State sample  $\{\xi^{k}_m\}_{m=1}^M$
                  \State  $x_m^{t, k+1} = x_m^{t, k} -\tau_1^k\left[\frac{1}{M}g_m(x_m^{t, k}; \xi_m^k)  + \zeta_m^t  \right]$ for $m=1,2 ,\dots, M$
            \EndFor
            \State $x_m^{t+1, 0} = x_m^{t, K}$ for $m=1,2,\dots, M$ 
            \State $\zeta_m^{t+1} = \zeta_m^t + \tau_2\left(x_m^{t+1, 0} - \sum_{j \in N_i}W_{i,j}x_j^{t+1,0} \right)$ for $m=1,2,\dots, M$
        \EndFor
        \State \textbf{Output:} $\bar x^T = \frac{1}{M}\sum_{m=1}^M x_m^{T, K}$ 
    \end{algorithmic} 
\label{alg:led}
\end{algorithm}

\begin{theorem}
Under Assumption~\ref{assum:global}, \ref{assume:W} and \ref{assum:sc}, we choose inner loop iterations $K$ large enough such that $\bE \|X^{t, K} - X^*(\lambda^t)\|^2 \leq \delta$, where $X^*(\lambda^t) = \argmin_{X} \cL(X, \lambda^t)$ is the optimal primal variable corresponding to $\lambda^t$. If we initialize $\zeta^0 = \lambda^0 = 0$, the output $\bar x^T$ of LED satisfies
\begin{align*}
 & \bE F^s(\bar x^T) - \min_x F(x)  \leq \frac{4ML^2}{\mu^2(1-\sigma_2)} \left(1- \frac{(1-\sigma_2 )\mu}{2L}\right)^T \|X_0 - X^*\|^2 + \frac{33L^3}{\mu^2 M(1-\sigma_2)^2} \delta.
\end{align*}

\end{theorem}

\begin{proof}
Note that 
\begin{align*}
     \fakeeq \bE \|X^{T, K} - X^*\|^2 & \leq 2\bE \left[\|X^{T, K}-X^*(\lambda^T)\|^2 + \|X^*(\lambda^T) - X^*\|^2\right]\\
    & \leq 2\delta + 2\bE\|X^*(\lambda^T) - X^*(\lambda^*)\|^2.
\end{align*}
Since $ \cL(X, \lambda)$ is $\frac{\mu}{M}$-strongly convex and the largest singular value of $U$ is upper bounded by $\sqrt{2}$, by Proposition \ref{prop:duality}, we have
$$
\|X^*(\lambda^T) - X^*(\lambda^*)\|^2 \leq \frac{2M^2}{\mu^2}\|\lambda^T - \lambda^*\|^2 .
$$
Therefore, 
\begin{equation*}
    \bE \|X^{T, K} - X^*\|^2 \leq 2\delta + \frac{4M^2}{\mu^2}\bE \|\lambda^T - \lambda^*\|^2.
\end{equation*}
By Jensen's inequality,
$$
\bE \|\bar x^T -x^*_s\|^2 \leq \frac{1}{M}\sum_{m=1}^M\bE\|x_m^{T, K} - x^*\|^2 \leq \frac{2\delta}{M} + \frac{4M}{\mu^2}\bE \|\lambda^T - \lambda^*\|^2.
$$
Because $F(x)$ is $L$-smooth, 
\begin{align} \nonumber
    \bE F(\bar x^T) - \min_x F(x) & \leq \frac{L}{2}\bE \|\bar x^T - x^*\|^2 \\ \nonumber
    & \leq \frac{L\delta}{M} + \frac{2ML}{\mu^2}\bE \|\lambda^T - \lambda^*\|^2  \\ \nonumber
    &\leq \frac{L\delta}{M} + \frac{2ML}{\mu^2} \left[\left(1- \frac{(1-\sigma_2 )\mu}{4L}\right)^T \|\lambda^0 - \lambda^*\|^2  + \frac{16L^2}{(1-\sigma_2)^2M^2}\delta \right] \\
    & \leq \frac{2ML}{\mu^2} \left(1- \frac{(1-\sigma_2 )\mu}{2L}\right)^T \|\lambda^0 - \lambda^*\|^2  + \frac{33L^3}{\mu^2 M(1-\sigma_2)^2} \delta.
\end{align}
Combined with (\ref{eq:initial-lambda-bd-decen}), 
\begin{align}
    \bE F(\bar x^T) - \min_x F(x) \leq \frac{2ML}{\mu^2} \left(1- \frac{(1-\sigma_2 )\mu}{2L}\right)^T \frac{2L\|X_0 - X^*\|^2}{1-\sigma_2}  + \frac{33L^3}{\mu^2 M(1-\sigma_2)^2} \delta,
\end{align}

\end{proof}

\subsection{Special Case: A Centralized Algorithm}
\label{sec:special_centralized}

We present the centralized version of Algorithm~\ref{alg:acc-multi-agda} in Algorithm~\ref{alg:cen-acc-multi-agda}, which can be derived by setting the weight matrix as follows:
\[
W = \frac{1}{M} \mathbf{1}\mathbf{1}^\top = \frac{1}{M} 
\begin{pmatrix}
1 & 1  & \cdots & 1 \\
1 & 1 & \cdots & 1 \\
\vdots & \vdots & \ddots & \vdots \\
1 & 1 & \cdots & 1
\end{pmatrix}.
\]

\paragraph{Comprison with Algorithm~\ref{alg:multi-agda} } When there is no momentum, i.e., $\beta = 0$, it can be considered a variant of Algorithm~\ref{alg:multi-agda}. The key difference is that in Algorithm~\ref{alg:multi-agda}, each client's local function is adjusted by an additional quadratic term, which is the result of the reformulated Lagrangian~(\ref{eq:lagrangian_cen}).

\paragraph{Comparison with Scaffnew/ProxSkip } When $\beta = 0$, the algorithm is nearly identical to Scaffnew \citep{mishchenko2022proxskip}, with the primary difference being that (a) Scaffnew uses a random number of inner-loop iterations, whereas we fix it to be $K$ in our approach; (b) the initial point of next inner loop in Scaffnew is set to be the averaged iterate, i.e.,  $x_m^{t+1, 0} = \bar x^t$. Therefore, we can interpret the correction term in Scaffnew as analogous to the dual variable $\zeta_m$ used in our approach.

\begin{algorithm}[t]  
    \caption{Centralized Accelerated Gradient Ascent Multi-Stochastic Gradient Descent}
    \begin{algorithmic}[1] 
      \State \textbf{Input:} initial point $x^0$
      and $\{\zeta_m^0\}_{m=1}^M$.
      \State \textbf{Initialize} $x_m^{0,0} = x^0$, $\widetilde \zeta^0_m = \zeta^0_m$ for $m=1, 2,\dots, M$
        \For{$t = 0,1,2,\dots, T$}
            \For{$k = 0,1,2,..., K-1$}
                 \State sample  $\{\xi^{k}_m\}_{m=1}^M$
                  \State  $x_m^{t, k+1} = x_m^{t, k} -\tau_1^k\left[\frac{1}{M}g_m(x_m^{t, k}; \xi_m^k)  + \widetilde\zeta_m^t  \right]$ for $m=1,2 ,\dots, M$
            \EndFor
            \State $x_m^{t+1, 0} = x_m^{t, K}$ for $m=1,2,\dots, M$  and $\bar x^t =  \frac{1}{M}\sum_{j=1}^M x_j^{t+1,0} $  
            \State $\zeta_m^{t+1} = \widetilde\zeta_m^t + \tau_2\left(x_m^{t, K} - \bar x^t \right)$ for $m=1,2,\dots, M$
            \State $\widetilde\zeta_m^{t+1} = \zeta_m^{t+1} + \beta\left(\zeta_m^{t+1} - \zeta_m^{t} \right)$ for $m=1,2,\dots, M$
        \EndFor
        \State \textbf{Output:} $\bar x^T = \frac{1}{M}\sum_{m=1}^M x_m^{T, K}$  
    \end{algorithmic} 
\label{alg:cen-acc-multi-agda}
\end{algorithm}


\end{document}